%% file: main.tex
\documentclass[letterpaper, paper,11pt]{AAS}

\usepackage{bm}
\usepackage{amsmath}
\usepackage{amsfonts}
\usepackage{subcaption}
\usepackage[colorlinks=true, pdfstartview=FitV, linkcolor=black, citecolor= black, urlcolor= black]{hyperref}
\usepackage{overcite}
\usepackage{footnpag}
\usepackage{graphicx}
\usepackage{wrapfig}
\usepackage{fancyhdr}

\fancypagestyle{firstpage}{
    \fancyhf{}                    
    \fancyhead[L]{%
        \footnotesize
        Presented at the 2026 AAS/AIAA Astrodynamics Specialist Conference,\\
        Whistler, BC, July 26--30, 2026.
    }
}

\PaperNumber{26-756}

\begin{document}

\title{Passively Safe Convex Guidance for Cislunar Rendezvous and Proximity Operations}

\author{Ian M. Down\thanks{Astrodynamics Engineer, Advanced Space, LLC, 1400 W 122nd Ave, Suite 200, Westminster, CO 80234}, Connor Plaks\thanks{Navigation Engineer, Advanced Space, LLC, 1400 W 122nd Ave, Suite 200, Westminster, CO 80234},
Matthew Bolliger\thanks{Senior Astrodynamics Engineer, Advanced Space, LLC, 1400 W 122nd Ave, Suite 200, Westminster, CO 80234}, Michael Caudill\thanks{Former Senior Navigation Engineer, Advanced Space, LLC, 1400 W 122nd Ave, Suite 200, Westminster, CO 80234}
}

\maketitle{}
\thispagestyle{firstpage}

\begin{abstract}
This paper presents purely convex programs for passively safe impulsive rendezvous and proximity operations in cislunar orbits. Approach, arrival, and abort maneuvers are all designed and validated in the context of maneuver execution error and navigation uncertainty, and formulated for efficient onboard execution in the autonomous scenario. The outlined methods form the baseline onboard guidance routines for NASA's CAPSTONE 02 mission planned to demonstrate autonomous rendezvous and proximity operations capabilities in the southern 9:2 synodic near rectilinear halo orbit. High fidelity closed loop Monte Carlo simulations using the planned relative navigation sensor suite and measurement cadence verify the intended maneuver design performance.
\end{abstract}

\input{01_intro}
\input{02_models}
\input{03_socp}
\input{04_rpo}
\input{06_examples}
\input{07_conclusion}

\bibliographystyle{AAS_publication}
\bibliography{references}
\input{10_appendix}

\end{document}

%% file: 01_intro.tex
\section{Introduction}

The emerging cislunar economy and NASA's Artemis program have identified Near Rectilinear Halo Orbits (NRHOs) as critical staging locations for lunar exploration and deep space missions. The Gateway space station, originally planned for a southern $L_2$ 9:2 synodic NRHO, was to serve as a multi-use outpost requiring frequent autonomous rendezvous and proximity operations (RPO) for assembly, resupply, and crew transfer missions. With NASA shifting focus away from Gateway and towards lunar surface permanence as part of the Moon Base program, the NRHO remains a key long-term technology demonstration orbit. Its high stability/low station-keeping requirements, breadth of celestial observation opportunities (deep-space, limb, and crater observations) and corresponding lighting conditions, and high degree of nonlinearity allow for a host of absolute and relative guidance and navigation algorithms to be regularly stress-tested on flight hardware. However, the cislunar environment presents unique challenges for RPO in particular that distinguish it from well-established techniques in near-circular low-Earth orbit (LEO). Highly nonlinear three-body dynamics, long orbital periods, communication delays with Earth, unstable trajectory characteristics, and extreme lighting conditions demand a paradigm shift toward fully autonomous, onboard guidance and navigation systems capable of ensuring passive safety under operational uncertainty.

On June 28, 2022, NASA and Advanced Space's Cislunar Autonomous Positioning System Technology Operations and Navigation Experiment (CAPSTONE) spacecraft launched into its low-energy lunar transfer, successfully inserting into a southern 9:2 synodic NRHO on November 13, 2022 \cite{cheetham2021cislunar,gardner2021capstone}. Since then, it has continued to demonstrate key enabling technologies that advance the utility and security of cislunar space, including autonomous cooperative positioning, navigation, and timing in the cislunar environment and characterization of the orbit's properties. After nearly four years of successful technology maturation, NASA’s activities on CAPSTONE concluded in June 2026. A follow-on mission, CAPSTONE 02, is planned to soon take its place in providing further critical technology demonstration and maturation in support of the Artemis program and NASA's Moon Base: humanity's first outpost on the lunar surface. Slated to launch in late 2027, the CAPSTONE 02 mission will send two small spacecraft to the 9:2 NRHO to demonstrate autonomous rendezvous and proximity operations, autonomous navigation, and cislunar communication capabilities, while also characterizing the radiation environment at the Moon. The mission aims to fill technology gaps critical for future lunar and cislunar infrastructure operations. This paper details the algorithms that will be used for autonomous real-time onboard maneuver planning during RPO demonstrations for the CAPSTONE 02 mission.

Traditional RPO techniques rely on iterative ground-in-the-loop planning and execution, leveraging nearly continuous ground-based tracking and pre-computed maneuver sequences that cannot adapt to real-time dispersions in navigation uncertainty or maneuver execution errors. In the autonomous scenario, onboard guidance algorithms must rapidly compute optimal maneuvers that guarantee passive safety: the statistical assurance that collision with the target is avoided even under complete loss of control, while accounting for execution errors and evolving state uncertainty. Convex optimization, and specifically Second Order Cone Programming (SOCP), offers a compelling solution to this challenge. Unlike nonlinear optimization methods that require initial guesses and may converge to local minima, convex programs possess deterministic convergence properties, bounded computation times, automatic feasibility verification, and global minima, making them very attractive in onboard, real-time operations. Successive convexification/sequential convex programming (SCP) take advantage of these properties to solve a series of convex sub-problems as part of an outer iteration loop that maintains a desired trust-region while updating linearizations of nonconvex functions. While empirically shown to be robust for a number of aerospace guidance problems, including rendezvous in an NRHO\cite{elango2025successive,elango2026continuous}, these algorithms still lack decisive numerical performance guarantees while at the same time introduce a steep increase in algorithmic complexity when compared to a single convex sub-problem. For particular problems with many nonlinear constraints, continuous control, and time appearing as a free variable, SCP is necessary to drive a solution. However, under the fixed-time premise, it is shown that simple, purely convex programs are sufficient for passively safe RPO in cislunar space, provided the relative motion is accurately captured by a first-order flow map.

Thus, this paper presents purely convex programs for designing passively safe impulsive rendezvous and proximity operations approach, arrival, and abort maneuvers in cislunar orbits. The methodology is not exclusive to cislunar trajectories, but near-circular two-body orbits likely benefit more by taking advantage of the exclusive reduced dynamical structure of that solution space. The method leverages a reduced but conservative maneuver execution error (MEE) model based on the Gates formulation, enabling uncertainty-aware trajectory design to be cast in second-order cone form. Chaser relative motion is modeled by the target's inertial state transition matrix generated from high-fidelity dynamics, while constraints are formulated in the target's Sun-local-vertical-local-horizontal (SLVLH) frame for fixed target visibility. Instantaneous safety constraints are incorporated by ellipsoid projection that enforce statistical separation from a target-centered keep out sphere. Approximated passive safety is evaluated on a rolling time-horizon over a discretized, regularized time grid. Anticipated violations are used to update instantaneous safety constraints in the maneuver design process. Terminal distributions are constrained to a discrete polyhedral docking cone whose angle is defined relative to the chaser's camera's field-of-view. Two high fidelity closed loop Monte Carlo simulations (one at apolune, one at perilune) using the planned relative navigation sensor suite and measurement cadence verify the intended maneuver design performance across the vast dynamic range of the NRHO. This paper is closely tied to its counterpart\cite{plaks2026closed}, which extensively details CAPSTONE 02 end-to-end closed-loop RPO performance analysis.

%% file: 02_models.tex
\section{Modeling}

\subsection{Sun-Local-Vertical-Local-Horizontal Frame}
\noindent
\begin{minipage}[t]{0.35\textwidth}
    \vspace{0pt}
    Let $\mathcal{S}$ denote the SLVLH frame\cite{gerstenmaier2019international}. Relative motion is constrained in this frame during maneuver design because the Sun geometry is fixed relative to the target, making motion planning relative to target visibility constraints during RPO activity convenient. Given target and Sun states in the inertial frame $\mathcal{I}$, the instantaneous SLVLH frame's construction and geometry are depicted to the right. Note that the axes are not congruent with the traditional two-body LVLH frame. A nominal rendezvous approach profile begins in the $+\hat{z}$, with the Sun on the chaser's back for optimal target visibility.
\end{minipage}%
\begin{minipage}[t]{0.65\textwidth}
    \vspace{0pt}
    \centering
    \includegraphics[width=0.7\textwidth]{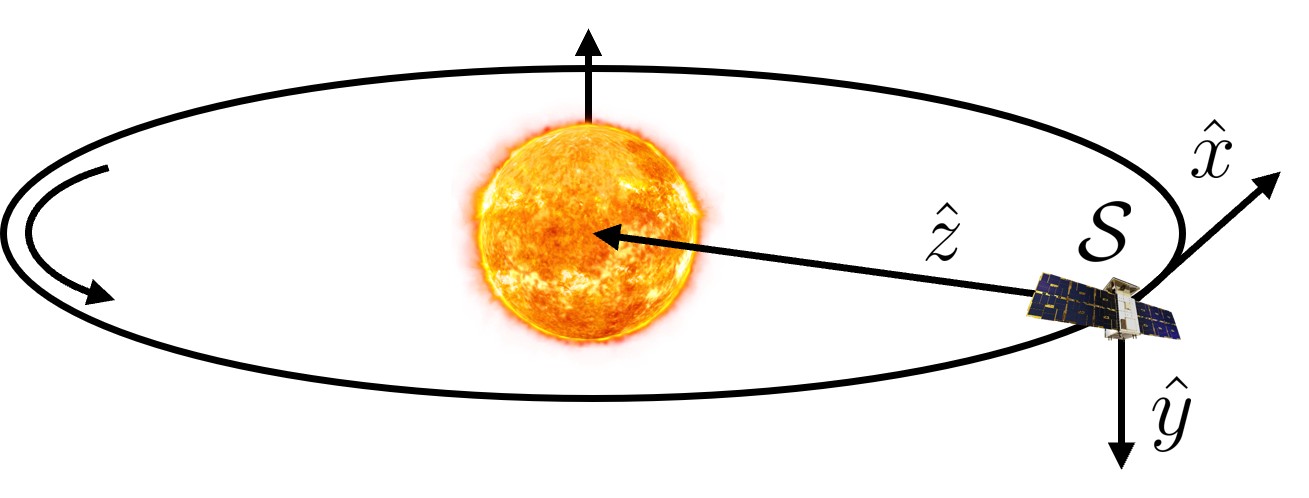}
    \begin{align}
        \Delta\boldsymbol{r} = \boldsymbol{r}_s - \boldsymbol{r}_t,\quad \Delta\boldsymbol{v} = \boldsymbol{v}_s - \boldsymbol{v}_t,\quad \Delta\boldsymbol{a} = \boldsymbol{a}_s - \boldsymbol{a}_t
    \end{align}
    \begin{align}
        \hat{\boldsymbol{z}} = \frac{\Delta\boldsymbol{r}}{||\Delta\boldsymbol{r}||},\quad \hat{\boldsymbol{y}}=-\frac{\Delta\boldsymbol{r}\times\Delta\boldsymbol{v}}{||\Delta\boldsymbol{r}\times\Delta\boldsymbol{v}||},\quad \hat{\boldsymbol{x}} = \hat{\boldsymbol{y}}\times\hat{\boldsymbol{z}}
    \end{align}
    \begin{align}
        \boldsymbol{R}_{\mathcal{I}}^{\mathcal{S}} = [\hat{\boldsymbol{x}},\:\hat{\boldsymbol{y}},\:\hat{\boldsymbol{z}}]^\text{T},\quad\dot{\boldsymbol{R}}_{\mathcal{I}}^{\mathcal{S}} =-\left[\boldsymbol{\omega}_{\mathcal{S}/\mathcal{I}}\right]_\times\boldsymbol{R}_{\mathcal{I}}^{\mathcal{S}}
    \end{align}
    \begin{gather}
        \boldsymbol{\omega}_{\mathcal{S}/\mathcal{I}}=-\frac{||\Delta\boldsymbol{r}\times\Delta\boldsymbol{v}||}{||\Delta\boldsymbol{r}||^2}\hat{\boldsymbol{y}} - \frac{||\Delta\boldsymbol{r}||(\Delta\boldsymbol{a}\cdot\hat{\boldsymbol{y}})}{||\Delta\boldsymbol{r}\times\Delta\boldsymbol{v}||}\hat{\boldsymbol{z}}
    \end{gather}
\end{minipage}

\subsection{Relative Motion}
Modeling relative motion around two-body dominated trajectories has a well-established base of literature. The autonomous linear Clohessy-Wiltshire-Hill (CWH) equations describe relative motion about a circular reference orbit, while the non-autonomous Tschauner-Hempel (TH) equations describe motion near an elliptical reference orbit. Furthermore, relative motion can be captured by first-order models of relative orbital element sets. Many adaptations, reductions, and solutions to these models including various perturbations have been published, studied, and leveraged\cite{sullivan2017comprehensive,down2024autonomous}. Relative motion in cislunar space lacks a low-level unified solution space because competing gravitational forces destroy the system's integrability. There is no general analytical solution to the circular restricted three-body problem, so there is no general analytical solution to relative motion in the circular restricted three-body problem (CR3BP). Still, nonlinear and linearized equations of relative motion for the CR3BP and ER3BP can be derived\cite{franzini2019relative,khoury2020orbital}, and modal decompositions of the linearized variational space about periodic reference trajectories aim to locally organize approximate solutions\cite{elliott2022describing,vela2026application}. One important difference between relative motion in LEO versus cislunar space is the period of curvature. The average LEO orbital period is around 110 minutes, meaning during a 24 hour long rendezvous sequence, a chaser is experiencing approximately 13 repeated cycles of motion. This leads to approach and circumferential trajectories like walking ellipses\cite{vavrina2019safe}. In contrast, many useful cislunar orbits have periods on the order of days to tens of days. The 9:2 synodic NRHO has a period of approximately 6.5 days, with both very fast and very slow dynamic regions. Around apolune, relative motion is effectively rectilinear over a 24 hour horizon\footnote{Rectilinear in the literal sense: straight-line, double-integrator-like dynamics.}. In contrast, relative motion can have very strong curvature over a 1 hour period near perilune, leading to strong relative state sensitivity. In this sense, rendezvous in longer period cislunar orbits is actually geometrically simpler than in LEO given the same baseline rendezvous timeline (i.e. an 8 hour crew day).

Because the CAPSTONE 02 spacecraft have effectively impulsive propulsion systems, and the maneuver design process exclusively involves fixed time-of-flight problems\footnote{This does not mean the RPO schedule onboard needs to be fixed. Trajectory arcs can still be delayed, shortened, or extended to accommodate fault management, navigation solution convergence, etc.}, the relative motion of the chaser is modeled by the target's state transition matrix (STM) as a linear flow map. This keeps the problem convex, while also allowing full, high-fidelity dynamics to be included in the model (to first-order). In addition, this is then a generic formulation that can be used agnostic to any dynamical regime/model as long as chaser motion occurs close enough to its target such that the first-order flow map accurately captures the relative motion over a given time horizon.

The CAPSTONE 02 navigation filter simultaneously estimates the dual inertial state of the two spacecraft\cite{plaks2026closed}. Let $\boldsymbol{x}=[\boldsymbol{x_t},\:\boldsymbol{x}_c]^\text{T}$ be the dual inertial state. Then the relative mean and covariance are
\begin{align}
    \delta\boldsymbol{m} &= \boldsymbol{x}_c - \boldsymbol{x}_t \\
    \delta\boldsymbol{P} &= \boldsymbol{P}_{cc} + \boldsymbol{P}_{tt} - \boldsymbol{P}_{ct} - \boldsymbol{P}_{tc}
\end{align}
The relative mean and covariance are mapped by the STM including an initial time maneuver as
\begin{align}
    \delta\boldsymbol{m}(t_f) &= \boldsymbol{\Phi}(t_f,t_0)\delta\boldsymbol{m}(t_0) + [\boldsymbol{\Phi}_{rv}(t_f,t_0),\quad\boldsymbol{\Phi}_{vv}(t_f,t_0)]^\text{T}\,\Delta\boldsymbol{V}(t_0)\\
    \delta\boldsymbol{P}(t_f) &= \boldsymbol{\Phi}(t_f,t_0)\delta\boldsymbol{P}^{+}(t_0)\left[\boldsymbol{\Phi}(t_f,t_0)\right]^\text{T}\\
    \delta\boldsymbol{P}^{+}(t_0)&=\delta\boldsymbol{P}^{-}(t_0) + \begin{bmatrix}
        \boldsymbol{0}_{3\times3} & \boldsymbol{0}_{3\times3} \\
        \boldsymbol{0}_{3\times3} & \boldsymbol{P}_{\epsilon}(\Delta\boldsymbol{V})
    \end{bmatrix}
\end{align}
where $\boldsymbol{P}_{\epsilon}$ is the maneuver execution inflation term. The final mean relative state and covariance can then be transformed to the rotating SLVLH frame as
\begin{align}
    \delta\boldsymbol{m}^\mathcal{S}(t_f) &= \mathcal{R}(t_f)\delta\boldsymbol{m}(t_f) \\
    \delta\boldsymbol{P}^\mathcal{S}(t_f) &= \mathcal{R}(t_f)\delta\boldsymbol{P}(t_f)\mathcal{R}(t_f)^\text{T} \\
    \mathcal{R} &=
    \begin{bmatrix}
        \boldsymbol{R}^\mathcal{S}_\mathcal{I} & \boldsymbol{0}_{3\times3} \\
        \smash{\dot{\boldsymbol{R}}}^\mathcal{S}_\mathcal{I} & \boldsymbol{R}^\mathcal{S}_\mathcal{I}
    \end{bmatrix}
\end{align}
The combined transformation through time and frame is affine, hence maintaining the convex relationship between final SLVLH mean/covariance and initial inertial maneuver vector, provided a convex maneuver execution error model is employed.

\begin{figure}[!t]
    \centering
    \begin{subfigure}{0.37\textwidth}
        \includegraphics[width=\textwidth]{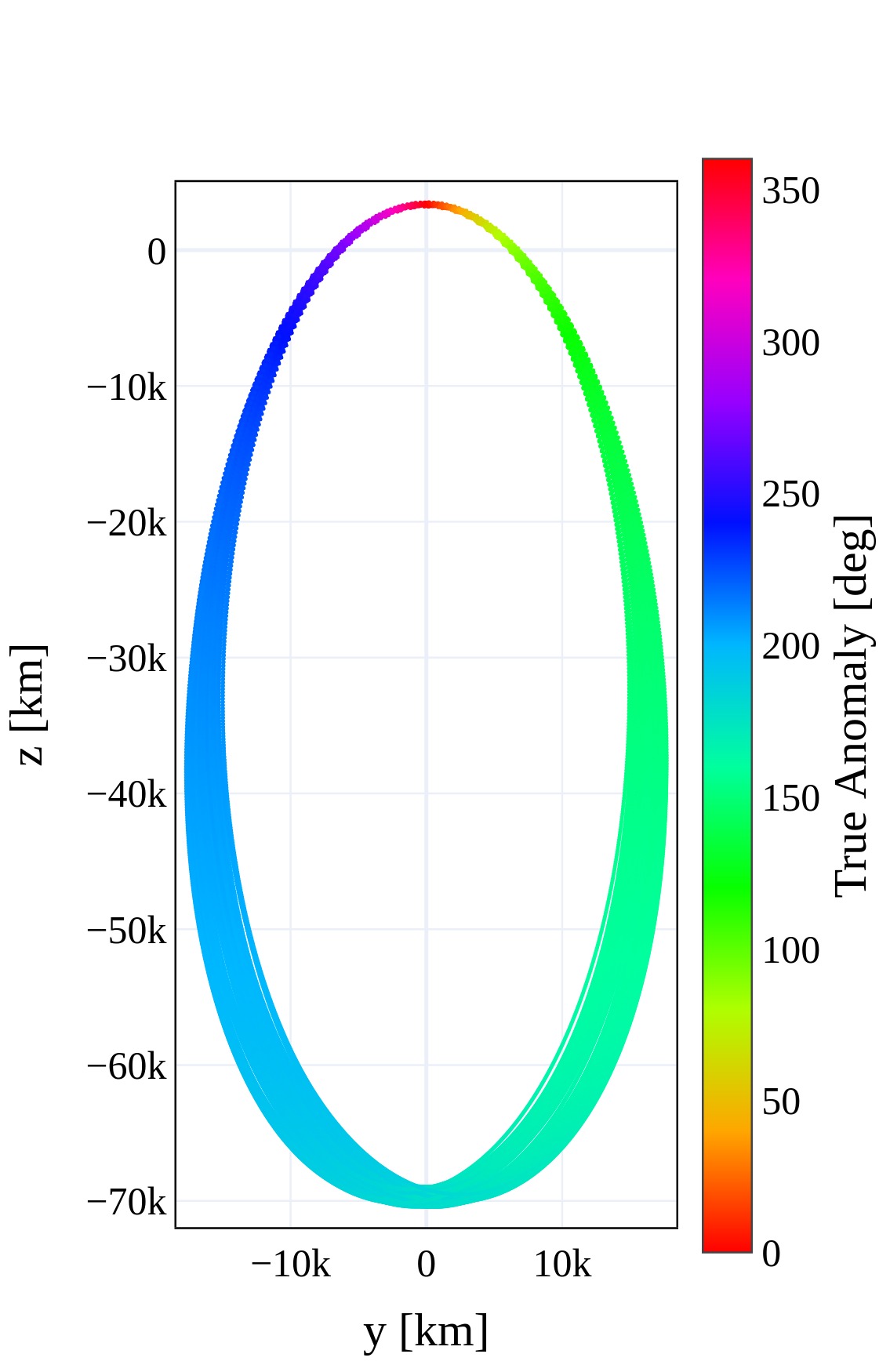}
        \caption{HFEM 9:2 synodic southern NRHO.\\}
    \end{subfigure}\hfill
    \begin{subfigure}{0.61\textwidth}
        \includegraphics[width=\textwidth]{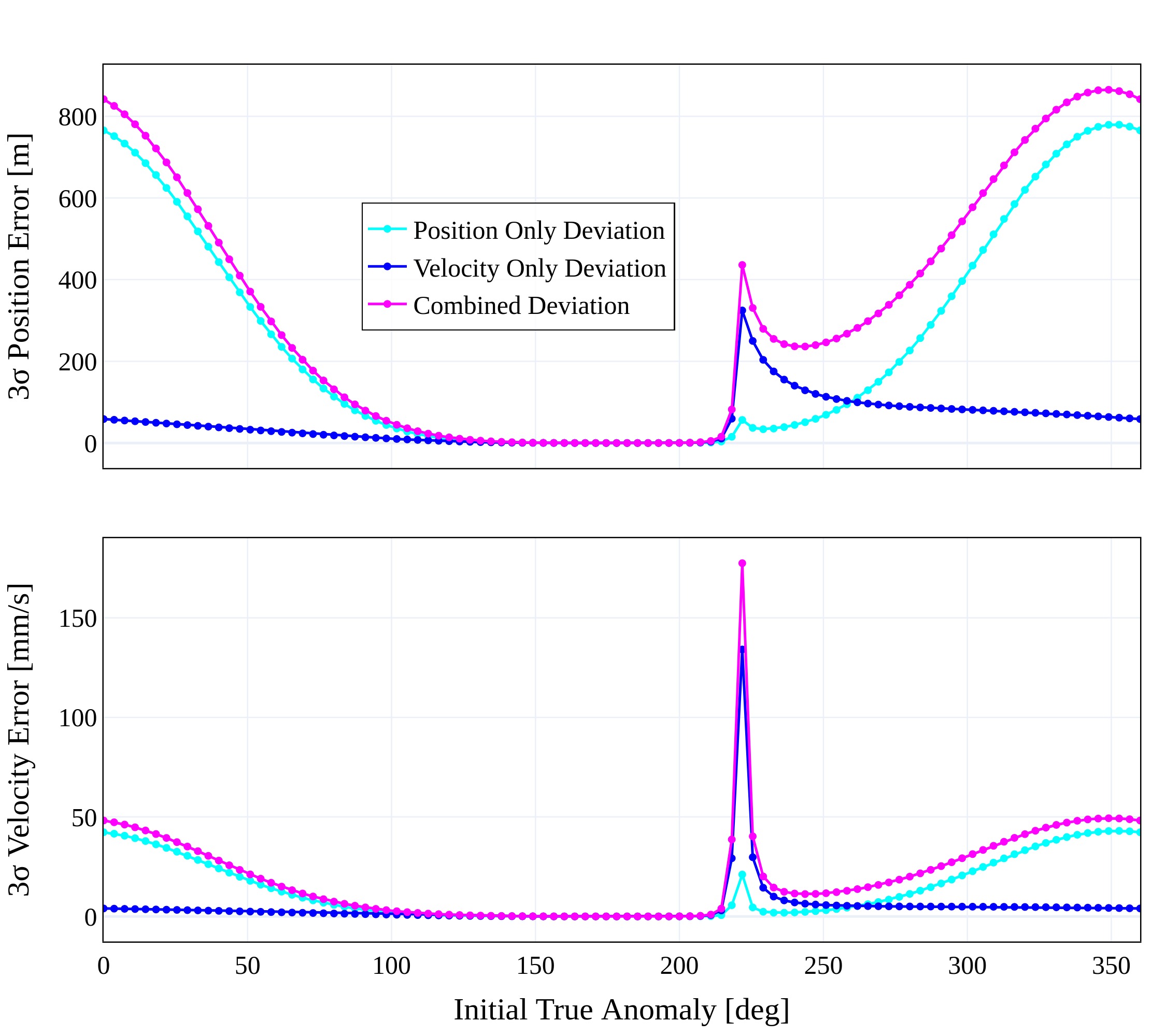}
        \caption{Monte Carlo results for 8 hour time horizon, 10 km \& 1 m/s initial deviation STM vs full dynamics mapping study.}
    \end{subfigure}\\
    \begin{subfigure}{0.61\textwidth}
        \includegraphics[width=\textwidth]{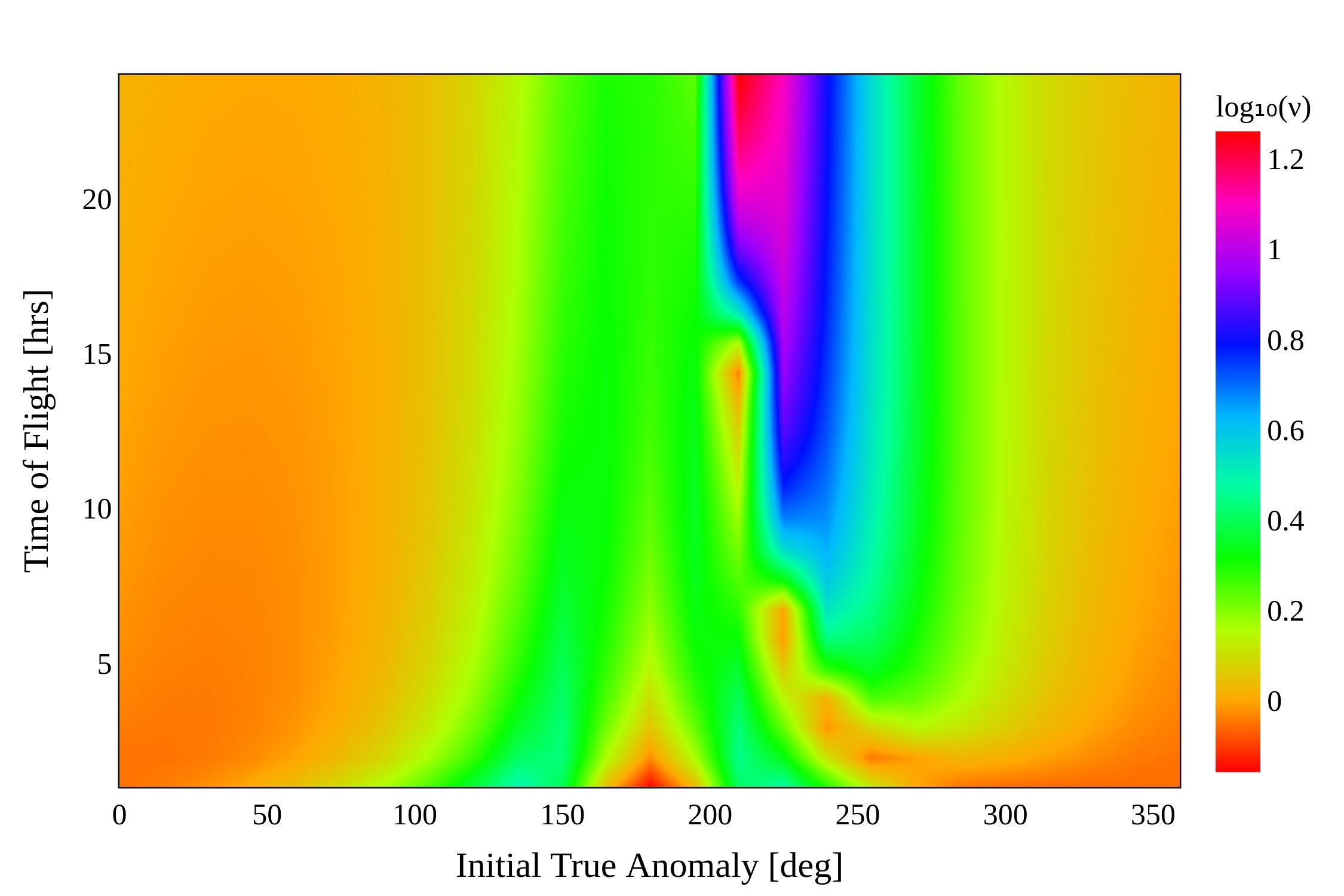}
        \caption{Nonlinear index results.}
    \end{subfigure}
    \caption{Relative motions studies for the CAPSTONE 02 NRHO.}
    \label{fig:relmotstudies}
\end{figure}

Two studies that examine the approximation capability of the inertial STM are now presented, with results shown in Fig. \ref{fig:relmotstudies}. 
The first study perturbs a reference state by points collected from a Fibonacci sphere, and propagates them over a time horizon, while also mapping them via the reference trajectory's STM. The final position and velocity magnitude error statistics are then computed to understand how well the STM approximates the flow map at given initial osculating true anomalies in the NRHO. Effects of deviating position and velocity independently are studied with $N=400$ Fibonacci distributions, while their combined effect is studied with a nested value of $N=40$. The 8 hour time horizon case with deviations of 10 km and 1 m/s is presented, with values chosen from expected nominal full onboard autonomy profiles. The second study computes a generalized nondimensional nonlinear index by comparing maximum first order stretching effects with second order maximum upper bounds on curvature effects\cite{kulik2025applications}. Both the first order STM $\boldsymbol{\Phi}$ and second order state transition tensor (STT) $\boldsymbol{\Psi}$ are propagated under Sun-Earth-Moon point-mass ephemeris dynamics along the NRHO reference trajectory over a time horizon. The final quantities are nondimensionalized by a characteristic length and time -- the CR3BP characteristic quantities are used here. The STT is then flattened to two dimensions such that the upper bound of the nonlinear index is
\begin{equation}
    \nu=\frac{\sqrt{\sigma_{\max}(\overline{\underline{\boldsymbol{\Psi}}})}}{\sigma_{\max}(\overline{\boldsymbol{\Phi}})}\label{eq:nonlinearindex}
\end{equation}
The radical of the STT multiplier is taken to account for the fact that the second order STT maps quadratic state deviations. The STM measures how much nearby trajectories separate, while the STT measures how rapidly that separation law changes from one nearby trajectory to another. Thus,
\begin{itemize}
    \item $\nu<1$: Stretching dominates curvature; dynamics are relatively linear.
    \item $\nu\approx1$: Stretching and curvature are equally important; nonlinear effects are significant.
    \item $\nu>1$: Curvature dominates stretching; the flow map changes rapidly with initial conditions, indicating strongly nonlinear dynamics.
\end{itemize}

The two studies examine the first order flow map in different ways. The nonlinear index measures the intrinsic curvature of the local flow map relative to its dominant linear stretching, whereas the Monte Carlo study measures the practical accuracy of the first-order STM over a finite uncertainty region. This is evident for the NRHO in Fig. \ref{fig:relmotstudies}, where the STM maps the motion extremely well for a majority of the orbit between true anomalies 100 and 200 degrees. That same region has a higher nonlinear index than perilune regions because the maximum linear stretching magnitude in that area is so small. Both studies reveal strong velocity dependent nonlinearity near true anomaly 220 degrees which is exploited for station-keeping\cite{davis2022orbit}.

\subsection{Maneuver Execution Error}
A primary driver in the CAPSTONE 02 safe rendezvous close approach distance is maneuver execution error (MEE)\cite{plaks2026closed}. The Gates' model for MEE accurately quantifies error with the incorporation of four error terms: both absolute and relative error associated with both the maneuver direction and off-axis components\cite{gates1963simplified}. The resulting instantaneous velocity covariance inflation in the body fixed maneuver frame $\mathcal{M}$ with maneuver direction in the $+z$-axis is
\begin{align}
    \boldsymbol{P}_{\epsilon}^\mathcal{M} &=
    \begin{bmatrix}
    \alpha^2 & 0 & 0 \\
    0 & \alpha^2 & 0 \\
    0 & 0 & \beta^2
    \end{bmatrix} \\
    \alpha^2 &= \sigma_{a,o}^2 + \sigma_{r,o}^2||\Delta\boldsymbol{V}||^2\\
    \beta^2 &= \sigma_{a,m}^2 + \sigma_{r,m}^2||\Delta\boldsymbol{V}||^2
\end{align} 
When expressed in any non-body-fixed frame, this MEE model becomes highly non-convex in its maneuver arguments because it requires the maneuver direction unit vector for conversion. In order to design uncertainty informed RPO maneuvers in the context of convex programming, a model reduction is required.

To simplify the Gates' model to a convex form, the maximum absolute standard deviation is applied isotropically to the distribution to remove dependence on maneuver direction in the absolute terms. This reduction is thus conservative, and any safety guarantees assured by a maneuver design solution are not lost when implemented by the physical system.
\begin{equation}
    \sigma_a=\max(\sigma_{a,o},\sigma_{a,m})
\end{equation}
The convex reduced model is more usefully expressed in the inertial frame as
\begin{align}
    \boldsymbol{P}_{\epsilon}(\Delta\boldsymbol{V}) = \sigma_{a}^2 \mathbb{I}_3 + \sigma_{r,m}^2[\Delta\boldsymbol{V}][\Delta\boldsymbol{V}]^\text{T} +\sigma_{r,o}^2[\Delta\boldsymbol{V}]_\times[\Delta\boldsymbol{V}]_\times^\text{T} \label{eq:reducedMEE}
\end{align}
where $[-]_\times$ is the left cross product matrix. This form will be leveraged in the construction of convex programs for RPO.

\subsection{Passive Safety}

The given proximity operations scenario is defined with a relative motion keep out sphere with radius $r_k$ centered on the target vehicle (at the origin in a relative motion reference frame), and time horizon bounds $t\in[t_0,\,t_f]$. To a confidence interval $\chi$, the chaser vehicle's statistical position $\delta\boldsymbol{r}$ relative to the target is found in the set described by the following uncertainty ellipsoid
\begin{align}
    \mathcal{E}(t;\chi):=\: [\delta\boldsymbol{r}(t)-\delta\boldsymbol{m}_r(t)]^\text{T}\delta\boldsymbol{P}_{rr}^{-1}(t)[\delta\boldsymbol{r}(t)-\delta\boldsymbol{m}_r(t)]\leq\chi^2\label{eq:ellipsoid}
\end{align}

Passive safety is said to be true if and only if the following expression holds over the time horizon in question:
\begin{align}
    \left[\min_{\delta\boldsymbol{r}(t)\in\mathcal{E}(t;\chi)}||\delta\boldsymbol{r}(t)||\right]>r_k\,\,\forall\,\, t\in[t_0,\,t_f]
    \label{eq:passive}
\end{align}

\begin{wrapfigure}{r}{0.45\textwidth}
    \centering
    \includegraphics[width=\linewidth]{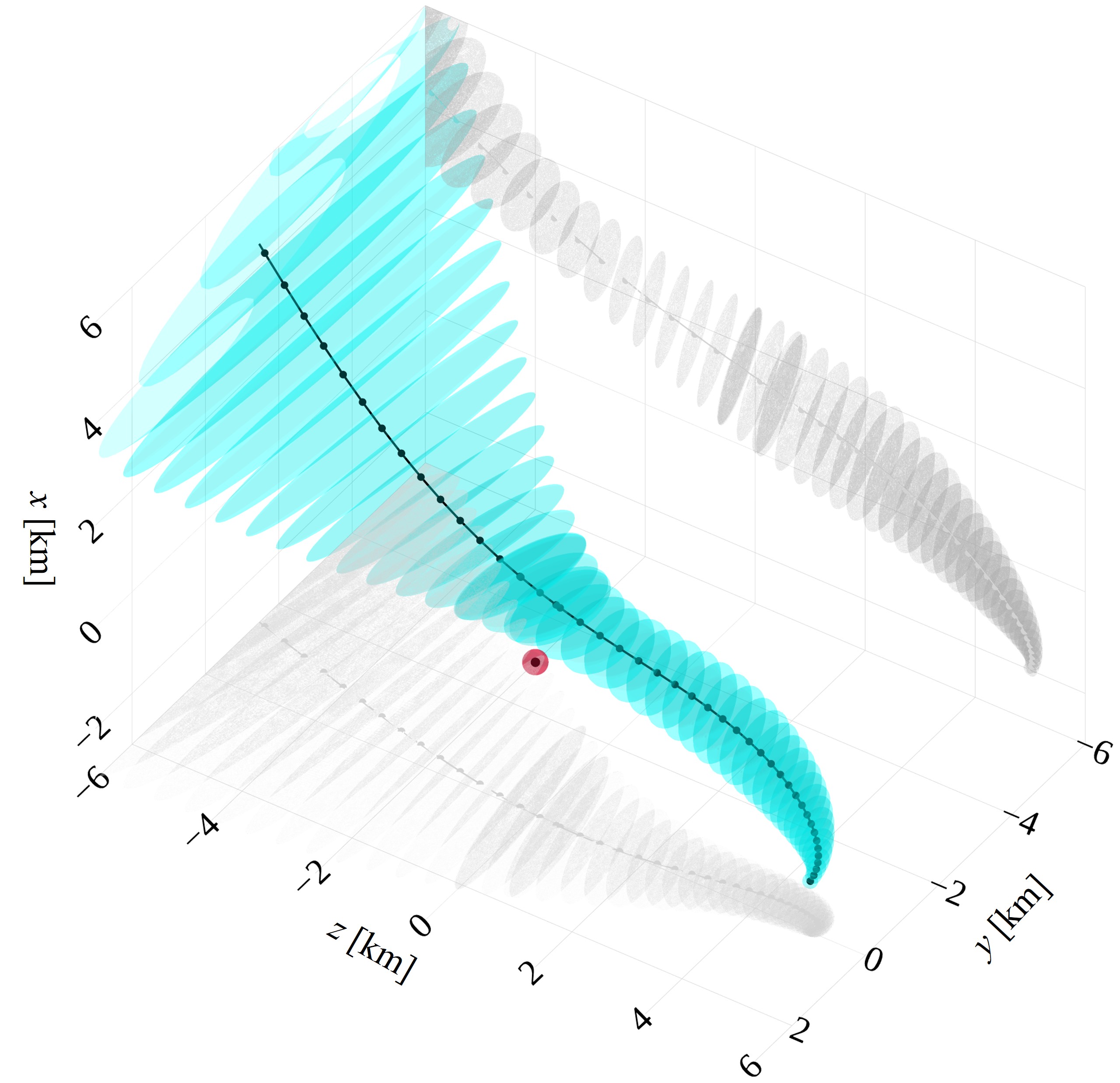}
    \caption{Discrete passive safety evaluation near perilune in the NRHO.}
    \label{fig:passivesafety}
\end{wrapfigure}
\noindent To verify passive safety, instantaneous safety is checked by solving Eq. \ref{eq:passive} at each node over a discretized time grid. For a particular instant in time, the expression is a convex constrained numerical optimization problem that nearly has a closed form solution. It may be reduced to a scalar nonlinear root finding problem over a strictly monotone function requiring a single eigen-decomposition. This is detailed in the Appendix. To ensure a more equal path-length distribution of discrete points, the discrete time grid of $N$ points is constructed as a uniform grid over regularized time\cite{leith2023time}. For the NRHO, the Sundman transform $\text{d}t = r\,\text{d}s$ is used, where $r$ is the target's distance from the Moon. Fictitious time $s$ is propagated alongside the target's trajectory to get $s_f$ over the passive safety time horizon. Then normal time $t$ is propagated alongside the target's trajectory with $s$ as the independent variable to retrieve the discrete times for instantaneous safety checks. For CAPSTONE 02, $\chi^2=11.345$ corresponding to 99\% confidence over a 3-dimensional distribution is used alongside a passive safety time horizon of 48 hours. An example of passive safety evaluation is shown in Fig. \ref{fig:passivesafety} starting at 250 degrees osculating true anomaly with a safety horizon of 6 hours for visualization.

%% file: 03_socp.tex
\section{Second Order Cone Programming}

Second order cone programming (SOCP) is a subset of convex programming \cite{boyd2004convex}. It contains both linear and quadratic programs. It is the lowest programming level that allows for constraints involving covariance to be written in convex form (as SOC constraints). Thus, all uncertainty informed RPO maneuver design strategies are reduced to this form. Formulating trajectory optimization problems into convex programs, and SOCPs specifically, is very attractive for autonomous onboard, real-time applications because these problems, if well-posed, have global optima, deterministic run times and iteration counts, and automatic parameter verification in the form of feasibility certificates. They also do not require an initial guess. In addition, several heavily tested real-time algorithms are readily available to solve this class of numerical optimization problems \cite{ECOS,goulart2024clarabel}.

\subsection{Forms}

\noindent
\begin{minipage}[t]{0.48\textwidth}
\vspace{0pt}
\subsubsection{Standard Form}
This form, along with its dual problem, are used by SOCP solvers\cite{ECOS}.
\begin{align}
    \min\quad & \boldsymbol{c}^\text{T}\boldsymbol{x} \\
    \text{s.t.}\quad & \boldsymbol{A}\boldsymbol{x} = \boldsymbol{b} \\
    & \boldsymbol{G}\boldsymbol{x} + \boldsymbol{s} = \boldsymbol{h},
    \quad \boldsymbol{s} \in \mathcal{K}
\end{align}
where $\boldsymbol{x}\in\mathbb{R}^n$ and $\boldsymbol{s}$ are the decision and slack variables, respectively, $\boldsymbol{b}\in\mathbb{R}^p$, $\boldsymbol{h}\in\mathbb{R}^M$, and $\mathcal{K}$ is the cone
\begin{gather}
    \mathcal{K} = \mathcal{Q}^{m_1} \times \cdots \times \mathcal{Q}^{m_N} \\
    \mathcal{Q}^{m} =
    \left\{
    (u_0,\boldsymbol{u}_1)\in\mathbb{R}\times\mathbb{R}^{m-1}
    \;:\;
    u_0 \geq \lVert \boldsymbol{u}_1 \rVert_2
    \right\}
\end{gather}
such that there are $p$ linear equality constraints, and $N$ SOC inequality constraints, each with dimension $m_i$, with $M=\sum_{i=1}^N m_i$.
\end{minipage}
\hfill
\begin{minipage}[t]{0.48\textwidth}
\vspace{0pt}
\subsubsection{Conic Form}
This form is more mathematically intuitive and is used in problem construction.
\begin{align}
    \min\quad & \boldsymbol{c}^\text{T}\boldsymbol{x} \\
    \text{s.t.}\quad & \boldsymbol{A}\boldsymbol{x} = \boldsymbol{b} \\
    & \lVert \boldsymbol{D}_i\boldsymbol{x} + \boldsymbol{e}_i \rVert_2
    \leq \boldsymbol{f}_i^\text{T}\boldsymbol{x} + g_i\\
    &\forall\, i=1,\ldots,N \nonumber
\end{align}
Once a problem is defined in conic form, it can be converted to standard form to solve. Namely, SOC problem data $\left\{\boldsymbol{D}_i,\boldsymbol{e}_i,\boldsymbol{f}_i,g_i\right\}$ must be converted to $\left\{\boldsymbol{G},\boldsymbol{h}\right\}$. Each SOC constraint is converted individually, and then stacked.
\begin{align}
    \boldsymbol{G}_i &=
    \begin{bmatrix}
        \boldsymbol{f}_i^\text{T} \\
        -\boldsymbol{D}_i
    \end{bmatrix},
    \,
    \boldsymbol{G} =
    \begin{bmatrix}
        \boldsymbol{G}_1^\text{T} &
        \cdots &
        \boldsymbol{G}_N^\text{T}
    \end{bmatrix}^\text{T}\\
    \boldsymbol{h}_i &=
    \begin{bmatrix}
        g_i \\
        -\boldsymbol{e}_i
    \end{bmatrix},
    \,
    \boldsymbol{h} =
    \begin{bmatrix}
        \boldsymbol{h}_1^\text{T} &
        \cdots &
        \boldsymbol{h}_N^\text{T}
    \end{bmatrix}^\text{T}
\end{align}
\end{minipage}

\subsection{Utilities}

\subsubsection{Epigraphs}
Both the standard and conic forms of the SOCP shown above contain linear objective functions. Convex quadratic and $\mathcal{L}_2$ norm cost functions are introduced into SOCPs through epigraphs \cite{rockafellar1997convex}. An epigraph of a function $f:\mathbb{R}^n\rightarrow\mathbb{R}$ is the set of all points lying on or above the graph of the function. That is
\begin{equation}
    \text{epi}(f)=\left\{(\boldsymbol{x},t)\in\mathbb{R}^n\times\mathbb{R},\:f(\boldsymbol{x})\leq t\right\}
\end{equation}
Importantly, a function is convex if and only if its epigraph is a convex set. By introducing an epigraph parameter $t$, a convex $\mathcal{L}_2$ cost function can be converted towards standard form with the addition of a SOC constraint:
\begin{equation}
\begin{array}{ccc}
\begin{aligned}
\min_{\boldsymbol{x}} \quad & ||\boldsymbol{x}||_2 \\
\text{s.t.} \quad & \boldsymbol{A}\boldsymbol{x}=\boldsymbol{b}\\ &
\end{aligned}
& \quad \longrightarrow \quad &
\begin{aligned}
\min_{\boldsymbol{x},t} \quad & t \\
\text{s.t.} \quad & \boldsymbol{A}\boldsymbol{x}=\boldsymbol{b} \\
& ||\boldsymbol{x}||_2 \leq t
\end{aligned}
\end{array}
\end{equation}

\subsubsection{Ellipsoid Projection}
\begin{figure}[htbp!]
    \centering
    \begin{minipage}{0.4\textwidth}
    \centering
    \includegraphics[width=0.6\textwidth]{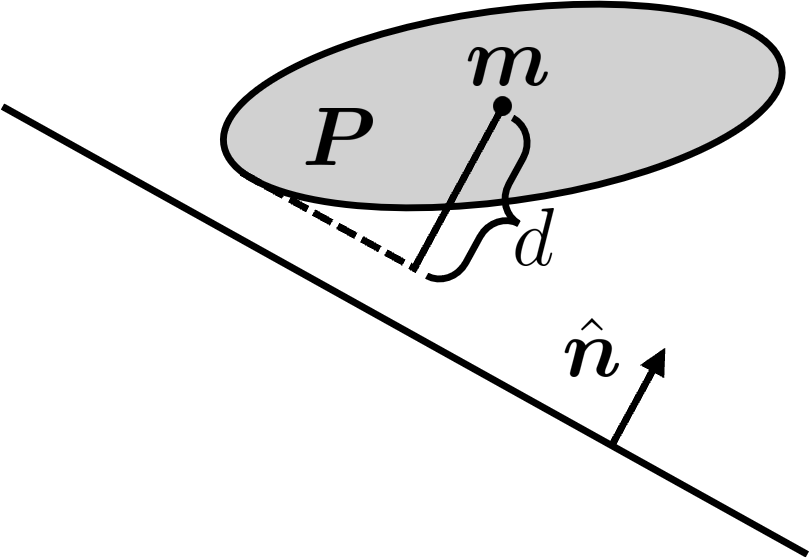}
    \caption{Ellipsoid projection.}
    \label{fig:ellipseProj}
    \end{minipage}%
    \begin{minipage}{0.4\textwidth}
    \begin{equation}
    \begin{aligned}
    d=\chi\sqrt{\hat{\boldsymbol{n}}^\text{T}\boldsymbol{P}\hat{\boldsymbol{n}}}
    \end{aligned}
    \end{equation}
    \end{minipage}
\end{figure}

Uncertainty in dynamical systems is often well captured and conveniently maintained by a covariance matrix $\boldsymbol{P}$. To properly ensure passive safety during an RPO sequence, convex chance constraints surrounding this matrix must be formulated in SOC form relative to maneuver inputs. The covariance constraints used in this work center around the concept of ellipsoid projection, as the exact instantaneous safety constraint detailed in the Appendix cannot be embedded directly within a convex program. A covariance ellipsoid centered on a mean state is defined by Eq. \ref{eq:ellipsoid}. For a given unit normal vector $\hat{\boldsymbol{n}}$, the projected covariance length $d$ along that direction over the specified confidence interval $\chi$ is given by Eq. \ref{fig:ellipseProj}3.
Position and velocity sub-blocks of a full state covariance matrix can be extracted, and constraints including the projected covariance in both position space and velocity space can be constructed independently. Using linear time and transformation mappings results in uncertainty projection constraints that are SOC convex in initial impulsive maneuver when the reduced MEE model of Eq. \ref{eq:reducedMEE} is used.

%% file: 04_rpo.tex
\section{Convex Guidance for RPO}
\begin{figure}[htpb!]
    \centering
    \includegraphics[width=0.7\linewidth]{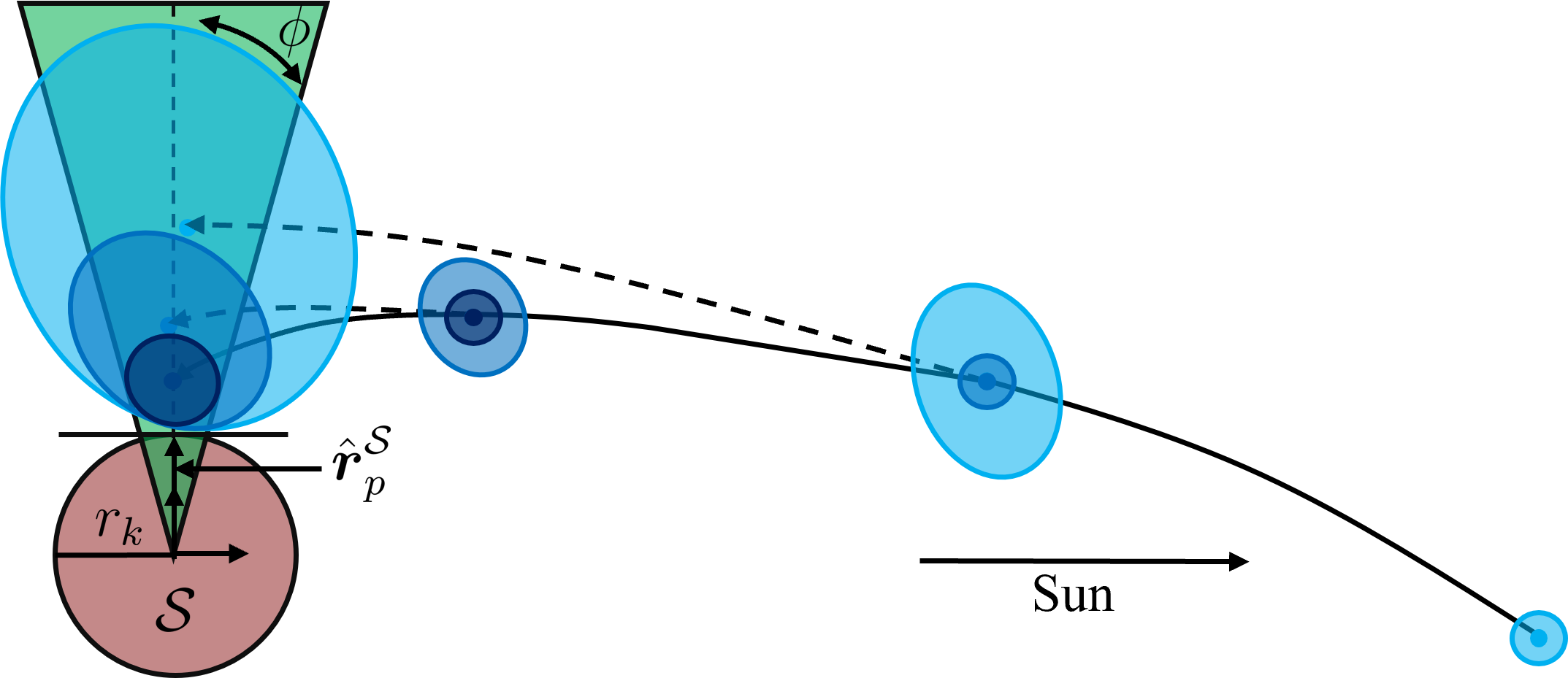}
    \caption{Uncertainty-informed rendezvous approach profile.}
    \label{fig:approach}
\end{figure}

Successive convexification/sequential convex programming (SCP) has proven to be a powerful tool for highly tailored trajectory design and optimization\cite{malyuta2022convex}. The methodology has been applied both in two-body dominated relative motion\cite{morgan2014model,berning2024chance}, as well as rendezvous in an NRHO\cite{elango2025successive,elango2026continuous}. These formulations lend themselves to nonconvex constraints, free time, and continuous control problems. Because the CAPSTONE 02 rendezvous problem involves fixed time horizons and impulsive control, while leveraging a discrete linear relative motion model and a conservatively convex maneuver execution error model, pure convex programming can be used for onboard maneuver design without the added complexity and eliminated compute guarantees of SCP. The following section outlines three guidance algorithms centered around the execution of convex programs: approach, arrival, and abort. The primary goal of the RPO activity is to minimize the chaser's arrival distance from the target's keep out sphere (KOS) while arriving inside a SLVLH docking cone and maintaining passive safety. This is done through a series of approach maneuvers as navigation uncertainty and time until arrival decrease. Once the chaser has arrived at the KOS, a maneuver is executed which nulls out mean relative velocity, with designed residual relative radial velocity ensuring free-drift safety under uncertainty. In the case of a fault scenario leading to the violation of passive safety, an active abort can be designed that statistically ensures the maneuver epoch is the linearized time of closest approach. A sketch of the approach profile is shown in Fig. \ref{fig:approach}.
Note that the below formulations treat SLVLH frame transformation matrices as deterministic quantities. Because the target's state is uncertain, that uncertainty maps into these matrices, which can be included into the formulation such that convexity holds. However, preliminary studies showed this uncertainty source was largely negligible, leading to arrival distance changes on the order of cm. Thus, for simplicity, these uncertainties are kept out of the formulation. These uncertainties do not affect passive safety evaluation, because that is executed in the inertial frame only.

\subsection{Approach}
Starting from as far away as 30 km, the chaser spacecraft executes a series of approach maneuvers that minimize the statistical close approach distance inside the docking cone at a predefined terminal epoch. The initial close rendezvous maneuver and all subsequent rendezvous correction maneuvers (RCMs) up until arrival leverage the same approach convex program, with the final RCM including additional constraints to constrain the terminal position distribution inside the docking cone itself\footnote{Including this constraint type in earlier RCMs leads to warped approach profiles because covariance growth downstream balloons from MEE.}. 
In between RCMs, relative navigation measurements (optical and crosslink) are collected, processed, and filtered. The number of RCMs and nominal RCM schedule are determined offline and uploaded to the spacecraft prior to RPO activity. The minimum time between RCMs is governed by the sum of time required for spacecraft bus slew and settle during measurement collection, number of measurements/navigation filter convergence, and burn prep prior to the next maneuver.

The design of a single approach maneuver follows a lightly iterative approach. First, the required STM and transformation matrix are computed, and the approach convex program is solved with a single safety constraint at the prescribed arrival time. The designed maneuver is then used to check passive safety under the full nonlinear dynamics and MEE parameters. If a violation occurs, an additional safety constraint is added at that discrete time and relative position direction. The approach convex program is re-run with any additional safety constraints necessary. If passive safety violation(s) is still detected, the absolute component of the reduced MEE model is updated to reflect the violation, and the entire loop repeats. For both apolune and perilune rendezvous trials, iteration is almost never required, but is included as a very simple, robust mechanism for handling the linear assumptions leveraged throughout the maneuver design process. The approach convex program and iterative algorithm are now mathematically detailed.

Let $\hat{\boldsymbol{r}}_p^{\mathcal{S}}$ and $\phi$ be the docking port axis and docking cone half-angle in the SLVLH frame as in Fig. \ref{fig:approach}, respectively. The base approach convex program minimizes the relative arrival distance while enforcing instantaneous safety at the arrival time along the docking port axis, keeping the mean arrival position within the docking cone itself. It is written as
\begin{align}
    \min_{\Delta\boldsymbol{V}(t_0)}\,\, &||\delta\boldsymbol{m}_r^{\mathcal{S}}(t_f)|| \label{eq:approach1}\\
    \text{s.t.}\,\,&||\delta\boldsymbol{m}_r^{\mathcal{S}}(t_f)||\cos\phi\leq(\hat{\boldsymbol{r}}_p^\mathcal{S})^\text{T}\,\delta\boldsymbol{m}_r^{\mathcal{S}}(t_f) \nonumber\\
    & r_k\leq(\hat{\boldsymbol{r}}_p^\mathcal{S})^\text{T}\,\delta\boldsymbol{m}_r^{\mathcal{S}}(t_f) - \chi\sqrt{(\hat{\boldsymbol{r}}_p^\mathcal{S})^\text{T}\delta\boldsymbol{P}_{rr}^{\mathcal{S}}(t_f)(\hat{\boldsymbol{r}}_p^\mathcal{S})}\nonumber
\end{align}
A regularized time grid is then constructed with two additional times included for passive safety evaluation. Both the prescribed arrival time, and the linear close approach time computed from the solution to Eq. \ref{eq:approach1} are included. Given the near straight-line relative motion around apolune in the NRHO, the mean linear close approach time and distance are highly effective at approximating the true nonlinear values, thus warranting an additional constraint check at this location. Closer to perilune, this particular time loses physical meaning, but including it in the general passive safety framework doesn't sacrifice performance in any way. The linear close approach time is computed as
\begin{equation}
    t_{lca} = t_f - \frac{\delta\boldsymbol{m}_r^\mathcal{S}(t_f)^\text{T}\delta\boldsymbol{m}_v^\mathcal{S}(t_f)}{||\delta\boldsymbol{m}_v^\mathcal{S}(t_f)||^2}
\end{equation}
where the mean relative quantities are mapped with the original STM and transformation matrix, and the $\Delta\boldsymbol{V}$ solution.

During passive safety evaluation of the above solution, let there be a violation at the $j$th discrete time point. Then the relative inertial position at $t_j$ from nonlinear propagation, $\delta\boldsymbol{m}_r^-(t_j)$ and the corresponding target STM $\boldsymbol{\Phi}(t_j,t_0)$ are used to construct the additional safety constraint
\begin{align}
    \delta\boldsymbol{m}(t_j) &= \boldsymbol{\Phi}(t_j,t_0)\delta\boldsymbol{m}(t_0) + [\boldsymbol{\Phi}_{rv}(t_j,t_0),\quad\boldsymbol{\Phi}_{vv}(t_j,t_0)]^\text{T}\,\Delta\boldsymbol{V}(t_0) \\
    r_k&\leq(\delta\hat{\boldsymbol{m}}_{r,j}^-)^\text{T}\,\delta\boldsymbol{m}_r(t_j) - \chi\sqrt{(\delta\hat{\boldsymbol{m}}_{r,j}^-)^\text{T}\delta\boldsymbol{P}_{rr}(t_j)(\delta\hat{\boldsymbol{m}}_{r,j}^-)}
\end{align}
These additional safety constraints are then added to the approach convex program, which is re-run for a refined solution. This waterfall approach to maneuver design allows for the incorporation of nonlinear trajectory information for passive safety constraints without requiring the full constraint linearization and iteration used in SCP. Thus, the maneuver design process isn't solely dependent on the STM as a relative motion model. This is appropriate under the assumption that the first solution is already close to feasible under nonlinear dynamics such that the addition of safety constraints won't dramatically change the solution geometry. This sidesteps the need to dynamically change the ellipsoid projection direction -- a primary driver for SCP formulations.

With an updated solution including additional passive safety constraints, passive safety is evaluated again, including the additional two discrete times. If any violations are found, a modification to the absolute component of the MEE model used in design is made to offset the worst nonlinear passive safety violation. Artificially inflating this value trades close approach distance for conservatism and design closure. Let $t_j$ be the time of the worst case safety violation, and $ d_j^*=||\delta\boldsymbol{r}^*_j|| - r_k<0$ be the distance violation at that time. Using the corresponding STM and violation vector $\delta\boldsymbol{r}^*_j$ (shown in the Appendix), a change in the absolute MEE value maps to the projected distance violation as
\begin{equation}
    | d_j^*|= \chi\sqrt{(\delta\hat{\boldsymbol{r}}_j^*)^\text{T}\delta\boldsymbol{P}_{rr}(t_j)\delta\hat{\boldsymbol{r}}_j^*}=\delta\sigma_a\cdot\chi\sqrt{(\delta\hat{\boldsymbol{r}}_j^*)^\text{T}\boldsymbol{\Phi}_{rv}(t_j,t_0)\boldsymbol{\Phi}_{rv}^\text{T}(t_j,t_0)\delta\hat{\boldsymbol{r}}_j^*}
\end{equation}
and thus the update law to artificially balance the violation is
\begin{align}
    \sigma_a^+&=\sigma_a^- + \delta\sigma_a \\
    \delta\sigma_a &= \frac{|d_j^*|}{\chi\sqrt{(\delta\hat{\boldsymbol{r}}_j^*)^\text{T}\boldsymbol{\Phi}_{rv}(t_j,t_0)\boldsymbol{\Phi}_{rv}^\text{T}(t_j,t_0)\delta\hat{\boldsymbol{r}}_j^*}}
\end{align}

Because the CAPSTONE 02 spacecraft have several different propulsion configurations with varying burn magnitude bands and thus Gates' model MEE parameters, an initialization and final check are required for the maneuver design process. To initialize the process, the maneuver required to move the initial mean position to $r_k$ along the docking port axis is computed analytically from the STM. After the maneuver design process is finished, the designed magnitude is checked against the initial band, and the process is restarted with the updated MEE parameters if a mismatch occurs. A block diagram of the approach maneuver design process is shown in Fig. \ref{fig:maneuverdesign} while the explicit conic form of the ellipsoid projection constraints are shown in the Appendix.
\begin{figure}[htpb!]
    \centering
    \includegraphics[width=0.9\linewidth]{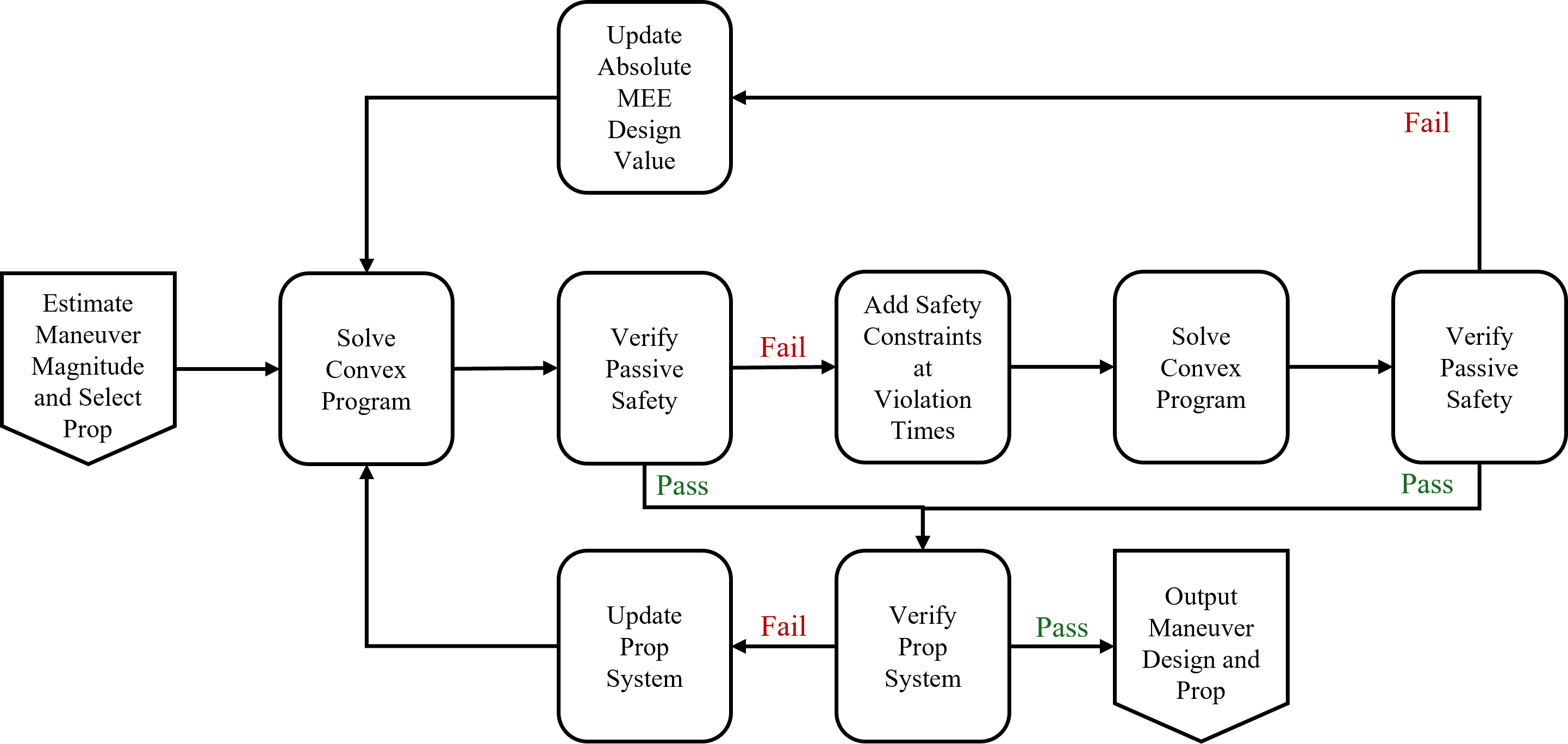}
    \caption{Flowchart of the onboard maneuver design process.}
    \label{fig:maneuverdesign}
\end{figure}

For the final RCM before arrival at the target's keep out sphere, additional constraints are added to constrain the terminal distribution to reside within the docking cone such that with 99\% confidence, the chaser will satisfy the docking cone constraint under uncertainty. To do this, the SOC is approximated by a polyhedral cone of $N$ sides, such that an additional $N$ ellipsoid projection constraints are appended to the original approach program. Let $\hat{\boldsymbol{e}}_3\equiv\hat{\boldsymbol{r}}_p^{\mathcal{S}}$. Then the polyhedral plane normal vectors $\hat{\boldsymbol{n}}$ are defined as
\begin{align}
    \hat{\boldsymbol{e}}_1 &= \frac{\hat{\boldsymbol{v}}_0 - (\hat{\boldsymbol{v}}_0\cdot\hat{\boldsymbol{e}}_3)\hat{\boldsymbol{e}}_3}{||\hat{\boldsymbol{v}}_0 - (\hat{\boldsymbol{v}}_0\cdot\hat{\boldsymbol{e}}_3)\hat{\boldsymbol{e}}_3||},\quad \hat{\boldsymbol{e}}_2 = \hat{\boldsymbol{e}}_3\times\hat{\boldsymbol{e}}_1,\quad\hat{\boldsymbol{v}}_0 = 
    \begin{cases}
        [1,0,0]^\text{T}, & \text{if } |\hat{\boldsymbol{e}}_3\cdot[1,0,0]^\text{T}|<0.9\\
        [0,1,0]^\text{T}, & \text{otherwise} 
    \end{cases} \\
    \hat{\boldsymbol{n}}(\theta) &= \frac{\hat{\boldsymbol{e}}_3 - (\hat{\boldsymbol{e}}_3\cdot\hat{\boldsymbol{v}}(\theta))\hat{\boldsymbol{v}}(\theta)}{||\hat{\boldsymbol{e}}_3 - (\hat{\boldsymbol{e}}_3\cdot\hat{\boldsymbol{v}}(\theta))\hat{\boldsymbol{v}}(\theta)||},\quad \hat{\boldsymbol{v}}(\theta)=\cos\phi\cdot\hat{\boldsymbol{e}}_3 + \sin\phi(\cos\theta\cdot\hat{\boldsymbol{e}}_1 + \sin\theta\cdot\hat{\boldsymbol{e}}_2)
\end{align}
The corresponding constraints are
\begin{align}
    0&\leq(\hat{\boldsymbol{n}}(\theta_i))^\text{T}\,\delta\boldsymbol{m}_r^{\mathcal{S}}(t_f) - \chi\sqrt{(\hat{\boldsymbol{n}}(\theta_i))^\text{T}\delta\boldsymbol{P}_{rr}^{\mathcal{S}}(t_f)\hat{\boldsymbol{n}}(\theta_i)}\\
    \theta_i &= \frac{2\pi(i-1)}{N},\quad i=1,...,N
\end{align}

\subsection{Arrival}
Once the chaser has arrived within the docking cone, an arrival maneuver is executed to null out the relative velocity to simulate the transition to a docking approach within the keep out sphere. To maintain passive safety after this maneuver in light of MEE, the arrival convex program minimizes the maneuver magnitude while constraining the post maneuver statistical velocity to be in the positive radial direction, and constraining the mean velocity to be parallel with the docking port direction. The mean velocity helps ensure the chaser doesn't exit the docking cone during post maneuver drift (under the straight-line relative motion assumption) to ensure consistency in lighting conditions for image capture upon arrival. The arrival convex program is mathematically detailed below, with time dropped as an argument.
\begin{align}
    \min_{\Delta\boldsymbol{V}}\,\, &||\Delta\boldsymbol{V}|| \label{eq:arrival}\\
    \text{s.t.}\,\,&\delta\boldsymbol{m}_v^{\mathcal{S},+} - \left[\left(\hat{\boldsymbol{r}}_p^\mathcal{S}\right)^\text{T}\delta\boldsymbol{m}_v^{\mathcal{S},+}\right]\hat{\boldsymbol{r}}_p^\mathcal{S}=\boldsymbol{0} \nonumber\\
    & 0\leq\left(\delta\hat{\boldsymbol{m}}_r^\mathcal{S}\right)^\text{T} \delta\boldsymbol{m}_v^{\mathcal{S},+} - \chi\sqrt{\left(\delta\hat{\boldsymbol{m}}_r^\mathcal{S}\right)^\text{T}\boldsymbol{P}_{vv}^{\mathcal{S},+}\delta\hat{\boldsymbol{m}}_r^\mathcal{S}}\nonumber
\end{align}
where $\delta\boldsymbol{m}_v^{\mathcal{S},+} = \delta\boldsymbol{m}_v^{\mathcal{S},-} + \boldsymbol{R}_{\mathcal{I}}^\mathcal{S}\Delta\boldsymbol{V}$. See the Appendix for an example on how the ellipsoid projection constraint is manipulated into conic form.

\subsection{Abort}
Lastly, in the event an abort is triggered and passive safety is violated, an active abort can be designed at any point along the approach profile by executing the arrival convex program without the mean velocity constraint. Under the straight-line relative motion assumption (apolune), this represents a deflection maneuver that makes the abort maneuver time the statistical closest approach distance. Executed near perilune, this abort strategy is still effective at creating a passively safe trajectory provided the waterfall strategy shown in Fig. \ref{fig:maneuverdesign} is applied.

%% file: 06_examples.tex
\section{Closed-Loop Simulation Examples}

Two closed-loop examples in the CAPSTONE 02 NRHO demonstrate the performance of the convex guidance algorithms introduced in this paper. Both cases use 100 Monte Carlo trials each to analyze closed-loop performance under uncertainty. A thorough explanation of these simulations are given by Plaks et al.\cite{plaks2026closed}. The first closed-loop simulation considers a 24-hour apolune rendezvous case officially beginning at the staging position in the SLVLH frame. This case is centered around near straight-line relative motion. The two vehicles begin in a 30-minute phased string-of-pearls formation, with one spacecraft designated as the ballistic target and the other the actively maneuvering chaser. A 200 meter KOS is enforced around the target. The chaser departs the formation at an osculating true anomaly of 160 degrees via a rendezvous start maneuver which targets a relative staging position in the SLVLH frame 24-hours downstream (designed on the ground). A clean-up maneuver is executed half-way to the staging position. The staging position is optimized offline for an near optimal approach profile for target illumination and dispersion collection. Once at the staging location, the 24-hour rendezvous approach begins. Onboard guidance is used to design the remaining rendezvous profile, which includes the target proximity maneuver (TPM), all subsequent RCMs, and the arrival insertion maneuver (AIM). One hour post-AIM, a ground designed departure initiation maneuver (DIM) brings the chaser back to the string-of-pearls formation. Table \ref{tab:MEE} summarizes the MEE parameters for two propulsion configurations used in the preliminary analysis of CAPSTONE 02 spacecraft RPO performance.
\begin{table}[htb]
	\fontsize{10}{10}\selectfont
    \caption{$\mathbf{1\sigma}$ Gates Model Maneuver Execution Error for Per Configuration}
   \label{tab:MEE}
        \centering 
   \begin{tabular}{|l | c | c | }
        \hline
        \hline 
        Thruster Configuration & 2x1 N Thrusters & 4x1 N Thrusters \\ \hline
        Burn Range (cm/sec)& 0 -- 45 & 45 -- 300 \\ \hline
        $\sigma_{a,m}$ (mm/sec)& 2.8 & 5.8 \\ \hline
        $\sigma_{r,m}$ ($\%$)& 2.1 & 1.1 \\ \hline
        $\sigma_{a,o}$ (mm/sec) & 0.0 & 0.0 \\ \hline
        $\sigma_{r,o}$ (rad) & 0.024 & 0.023 \\ \hline
      \hline
   \end{tabular}
\end{table}

\begin{figure}
    \centering
    \includegraphics[width=1\linewidth]{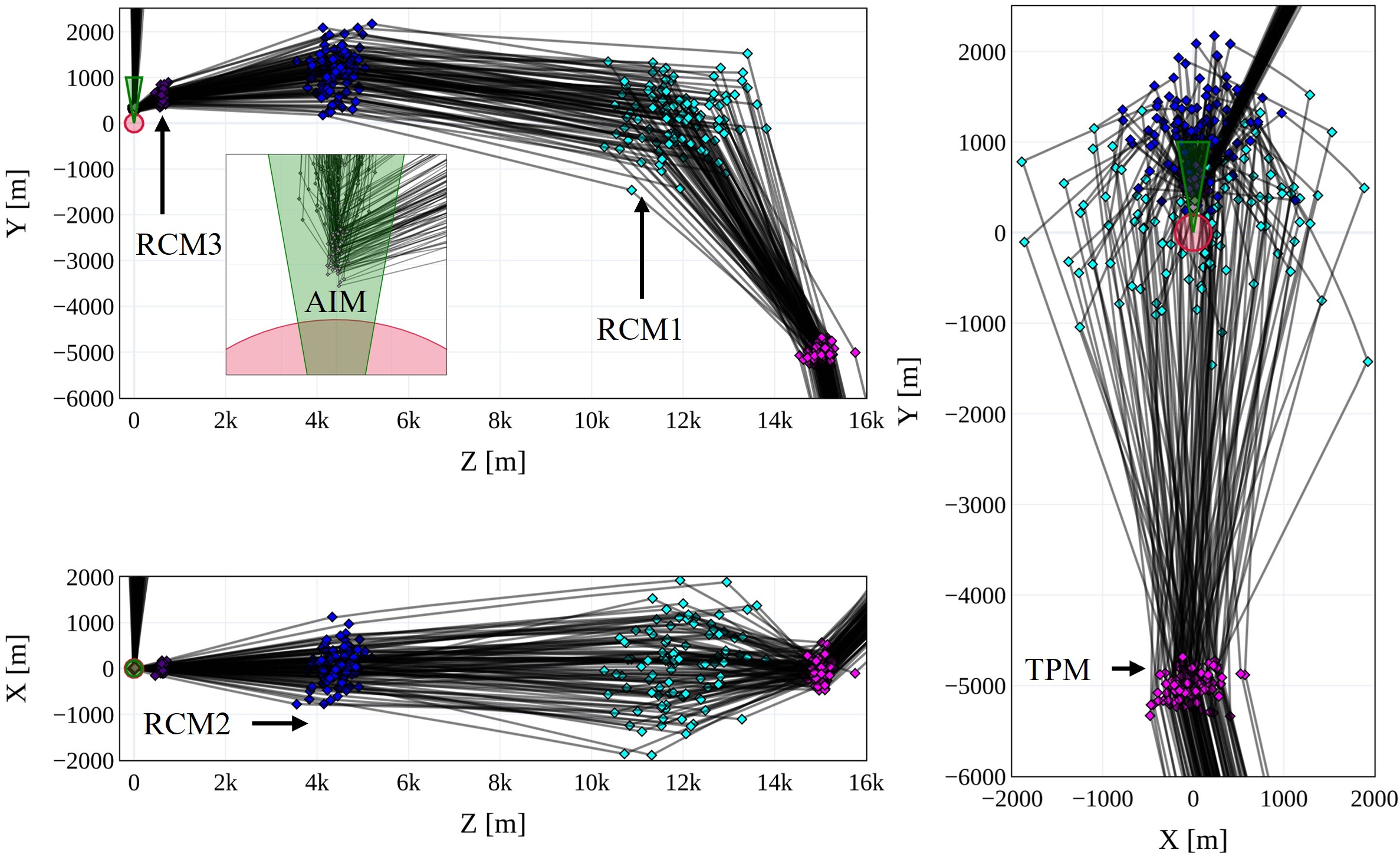}
    \caption{Trajectory dispersions for 24-hour, 3 RCM, rendezvous near apolune.}
    \label{fig:apolune_trajs}
\end{figure}

Proximity operations navigation in the simulation is closed-loop and measurement-driven, using ground two-way X-band range and Doppler prior to close operations, inter-spacecraft two-way S-band crosslink range and Doppler once the vehicles are within approximately 400 km, and optical bearing measurements inside roughly 100 km. Optical and crosslink tracks are modeled conservatively with a 50\% duty cycle; the resulting performance reflects the combined impact of passive safety constraints, MEE dispersion, navigation uncertainty, and post-maneuver reacquisition requirements. Each call to the guidance algorithms detailed above executes in sub-second compute time using a \textsc{Python} implementation, with significant speed-up expected in compiled flight software. This highlights the light-weight, deterministic footprint of conic solvers. The trajectory dispersion results are shown in Fig. \ref{fig:apolune_trajs}.
The results exemplify constraint satisfaction. Because of the projected uncertainty growth driven by MEE over the long 24-hour timeline, three RCMs are required for a meaningful close approach outside the KOS. The RCM number and nominal placement in time are determined offline using results from simulations like these. Between maneuvers, the straight-line relative motion characteristic of slow apolune dynamics is clearly prevalent.

\begin{figure}
    \centering
    \includegraphics[width=1\linewidth]{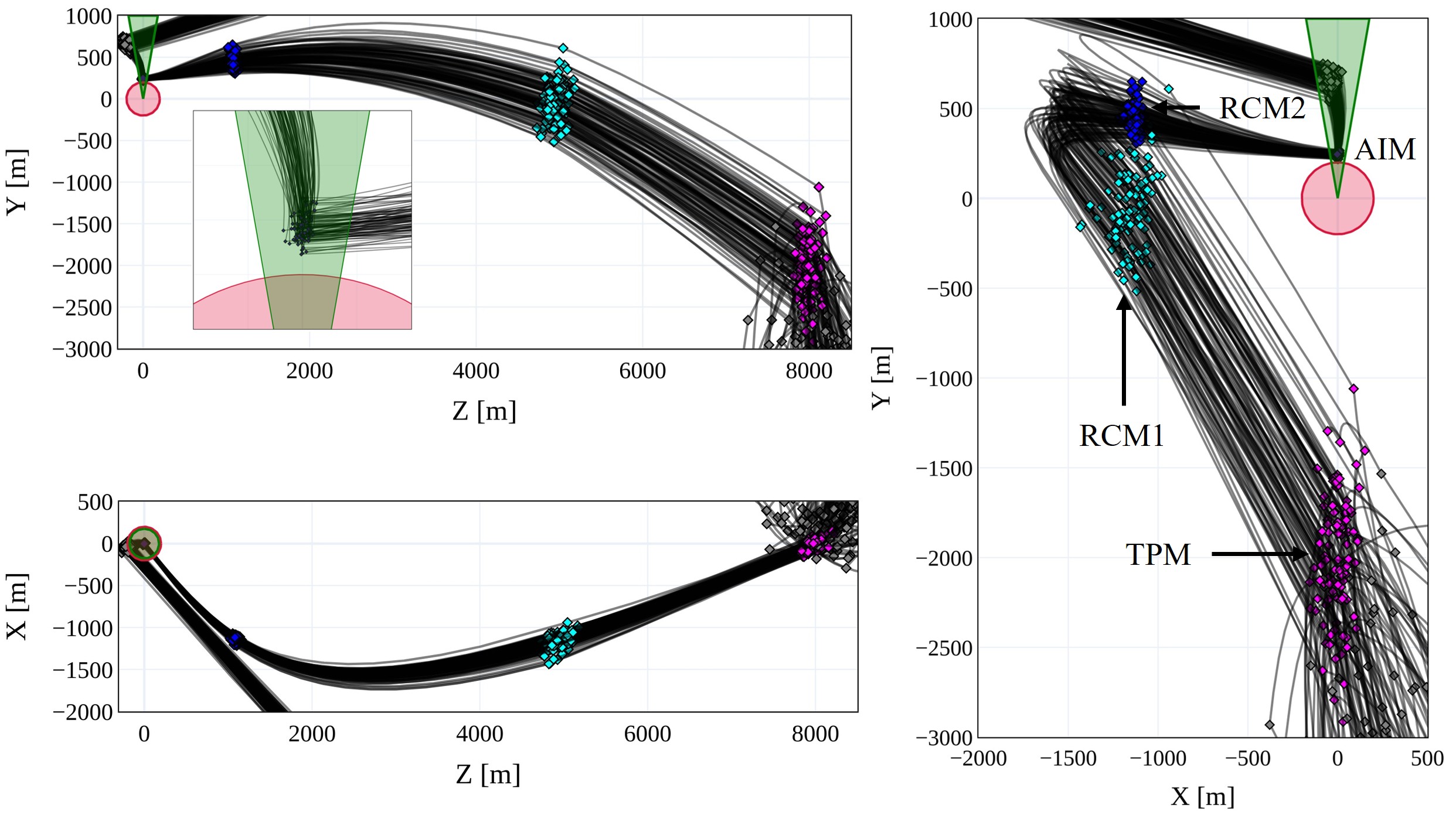}
    \caption{Trajectory dispersions for 8-hour, 2 RCM, rendezvous near perilune.}
    \label{fig:perilune_trajs}
\end{figure}

The second closed-loop simulation considers an 8-hour perilune rendezvous case designed to evaluate the passively safe convex guidance architecture in the strongly nonlinear portion of the CAPSTONE 02 NRHO. The ground-designed string-of-pearls to SLVLH staging maneuvers in this case are more complex in efforts to reduce overall system fuel expenditure and maximize system lifespan. They are detailed in the companion paper\cite{plaks2026closed}. TPM occurs 8 hours prior to perilune pass, meaning the arrival time is scheduled to occur at perilune. Once a target loiter is complete, autonomy is initiated, and onboard guidance brings the chaser into the target for approach and arrival. After the autonomy handover, the simulation uses only the relative navigation sensor suite, namely S-band crosslink range/Doppler and optical bearing measurements acquired in tandem at a conservative 50\% duty cycle. Onboard orbit determination and maneuver design are executed with a 20-minute processing assumption plus a 10-minute slew and burn lockout, yielding a 30-minute data-cutoff-to-execution timeliness for the autonomous approach. The case therefore highlights how the guidance algorithm performs when passive safety, navigation uncertainty, and maneuver execution error are all carried through a full closed-loop perilune approach. The trajectory dispersion results are shown in Fig. \ref{fig:perilune_trajs}.
Again, the chaser arrives at the KOS with safe constraint satisfaction. This example in particular highlights the convex guidance schemes' ability to handle highly nonlinear regimes using the underlying assumptions levied upon them to keep the algorithms convex. The chaser in this case arrives within the docking cone under 200 meters away from the KOS in all cases with only two RCMs -- an artifact of a much shorter approach horizon.

\begin{figure}[t]
    \centering
    \begin{subfigure}{0.35\textwidth}
        \includegraphics[width=\textwidth]{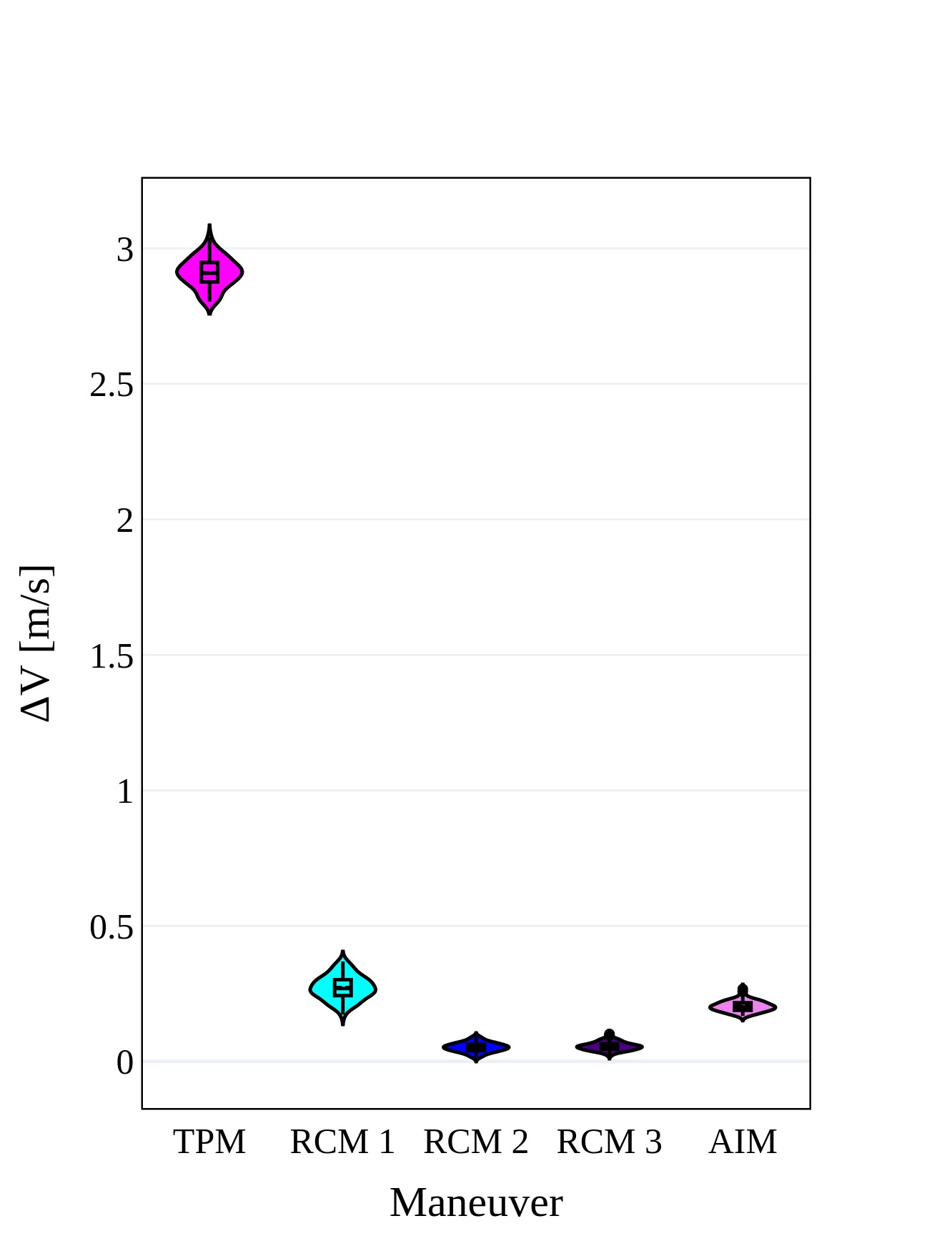}
        \caption{Apolune example.}
    \end{subfigure}
    \begin{subfigure}{0.35\textwidth}
        \includegraphics[width=\textwidth]{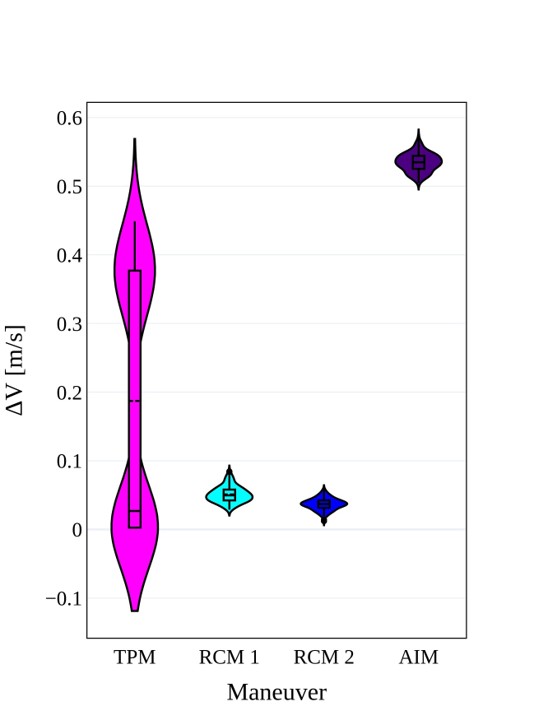}
        \caption{Perilune example.}
    \end{subfigure}
    \caption{Maneuver statistics during autonomous rendezvous using convex guidance schemes.}
    \label{fig:manstats}
\end{figure}

The maneuver statistics for both closed-loop examples are shown Fig. \ref{fig:manstats}. TPM in the apolune case is much larger than perilune because there are no natural dynamics to leverage in the apolune region. So, while the ballistic relative motion is easier to predict, it is also in general more costly to control. In addition, the far-field rendezvous maneuvers and subsequent relative velocity at TPM play a large role in its magnitude for both cases, which are detailed by Plaks et al.\cite{plaks2026closed}. The negative TPM magnitude values for the perilune case are solely a product of the statistical violin plot construction, and are, for obvious reasons, not real solutions. For some trials in both the apolune and perilune examples, RCM maneuver magnitudes fall very close to zero. Despite Table \ref{tab:MEE} reaching down to that value, maneuvers below a notional 1 cm/s threshold would be canceled in operations.

%% file: 07_conclusion.tex
\section{Conclusion}
This paper presented a purely convex guidance architecture for passively safe impulsive rendezvous and proximity operations in cislunar space, with specific application to the CAPSTONE 02 mission in a southern 9:2 synodic NRHO. By combining a first-order STM-based relative motion model, a conservative convex reduction of the Gates maneuver execution error model, and ellipsoid-projection safety constraints formulated as second-order cone constraints, the approach enables onboard design of approach, arrival, and abort maneuvers without the algorithmic complexity or convergence uncertainty of successive convexification. The resulting guidance framework preserves the key benefits of convex optimization for autonomy-critical applications, including deterministic convergence behavior, feasibility certification, and bounded computation time.

The proposed methods were shown to support the full close-operations sequence: long-horizon approach design toward the keep out sphere, terminal arrival with uncertainty-aware velocity shaping, and active abort generation when passive safety is threatened. The waterfall-style constraint update strategy further allowed nonlinear safety information from high-fidelity propagation to be incorporated while retaining a purely convex maneuver design core. In this way, the formulation remains computationally lightweight while still accommodating the strong variation in dynamical character encountered across the NRHO.

High-fidelity closed-loop Monte Carlo simulations verified the performance of the guidance architecture in both slow, near-rectilinear apolune conditions and strongly nonlinear perilune conditions using the planned CAPSTONE 02 relative navigation sensor suite and measurement cadence. In both cases, the chaser was guided to the target keep out sphere while satisfying passive safety and docking geometry constraints under maneuver execution error and navigation uncertainty. The results therefore demonstrate that purely convex, uncertainty-aware impulsive guidance is a practical and effective solution for autonomous cislunar RPO.

These methods form the baseline onboard guidance routines for CAPSTONE 02 and provide a scalable foundation for future autonomous proximity operations in NRHOs and other multi-body regimes. Future work will focus on continued flight software maturation, operational tuning with the integrated GNC architecture, and on-orbit demonstration of these capabilities in support of sustained lunar and cislunar infrastructure.

%% file: 10_appendix.tex
\appendix


\section*{Appendix: Instantaneous Safety Derivation}
The zero step for verifying instantaneous safety simply tests the covariance ellipsoid inequality at the point $\boldsymbol{r}=\boldsymbol{0}$. Provided this point is not contained in the ellipsoid, the following convex program and its solution may be applied to determine instantaneous safety.

To evaluate the inequality constraint present in the passive safety definition of Eq. \ref{eq:passive} for a single point in time, the following convex program must be solved. Time is dropped as an argument as well as $\delta$ notation for convenience.
\begin{equation}
    \min_{\boldsymbol{r}\in\mathcal{E}(\chi)}||\boldsymbol{r}|| \equiv \min_{\boldsymbol{r}\in\mathcal{E}(\chi)}\boldsymbol{r}^\text{T}\boldsymbol{r}
\end{equation}
The minimizer of both the norm and quadratic forms are the same because the square function is monotone on positive real axis. This is a consequence of the convex ellipsoid curvature as the admissible set and the recognition that the solution to both of these problems lie on the ellipsoid surface.

The goal is to obtain an analytical solution to this problem that may be efficiently evaluated at many discrete time points to declare passive safety. To begin this process, the Lagrangian is formed
\begin{equation}
    \mathcal{L} = \boldsymbol{r}^\text{T}\boldsymbol{r} + \lambda\left([\boldsymbol{r}-\boldsymbol{m}_r]^\text{T}\boldsymbol{P}_{rr}^{-1}[\boldsymbol{r}-\boldsymbol{m}_r]-\chi^2\right)
\end{equation}
\begin{align}
    \frac{\partial \mathcal{L}}{\partial\boldsymbol{r}}=\boldsymbol{0}\implies\boldsymbol{r}&=\left(\mathbb{I}+\lambda\boldsymbol{P}_{rr}^{-1}\right)^{-1}\lambda\boldsymbol{P}_{rr}^{-1}\boldsymbol{m}_r \\
    \boldsymbol{r}&=\boldsymbol{m}_r - \left(\mathbb{I}+\lambda\boldsymbol{P}_{rr}^{-1}\right)^{-1}\boldsymbol{m}_r \\
    &\text{because } \lambda>0,\:\boldsymbol{P}_{rr}\succ0
\end{align}
Now with some rearrangement, the first order optimality condition can be substituted into the ellipsoid equality constraint.
\begin{align}
    \boldsymbol{r}-\boldsymbol{m}_r=- \left(\mathbb{I}+\lambda\boldsymbol{P}_{rr}^{-1}\right)^{-1}\boldsymbol{m}_r\label{eq:subsol}\\
    \left[\left(\mathbb{I}+\lambda\boldsymbol{P}_{rr}^{-1}\right)^{-1}\boldsymbol{m}_r\right]^\text{T}\boldsymbol{P}_{rr}^{-1}\left[\left(\mathbb{I}+\lambda\boldsymbol{P}_{rr}^{-1}\right)^{-1}\boldsymbol{m}_r\right]=\chi^2 \label{eq:subconstraint}
\end{align}
Solving for $\lambda$ in Eq. \ref{eq:subconstraint} and using it in Eq. \ref{eq:subsol} completes this problem. A quest for $\lambda$ is now evoked.

As an input to this problem, we are working with the covariance matrix itself, not its inverse (the information matrix). To further simplify the problem, the covariance matrix must be diagonalized.
\begin{align}
    \boldsymbol{P}_{rr}=\boldsymbol{Q}\boldsymbol{\Sigma}_{rr}\boldsymbol{Q}^\text{T}&\implies\boldsymbol{P}_{rr}^{-1}=\boldsymbol{Q}\boldsymbol{\Sigma}_{rr}^{-1}\boldsymbol{Q}^\text{T}\\
    &\implies\left(\mathbb{I}+\lambda\boldsymbol{P}_{rr}^{-1}\right)^{-1} = \boldsymbol{Q}\left(\mathbb{I}+\lambda\boldsymbol{\Sigma}_{rr}^{-1}\right)^{-1}\boldsymbol{Q}^\text{T}
\end{align}
This matrix inverse relationship is only possible because the covariance is assumed positive definite. With the computation of $\boldsymbol{\Sigma}_{rr}$ from the eigen-decomposition of $\boldsymbol{P}_{rr}$, $\boldsymbol{\Sigma}_{rr}^{-1}$ is trivially computed as the element-wise inverse.

Let the mean position projected into the principal basis be
\begin{equation}
    \tilde{\boldsymbol{m}}_r=\boldsymbol{Q}^\text{T}\boldsymbol{m}_r
\end{equation}
then the equality constraint is now
\begin{equation}
    \tilde{\boldsymbol{m}}_r^\text{T}\left(\mathbb{I}+\lambda\boldsymbol{\Sigma}_{rr}^{-1}\right)^{-1}\boldsymbol{\Sigma}_{rr}^{-1}\left(\mathbb{I}+\lambda\boldsymbol{\Sigma}_{rr}^{-1}\right)^{-1}\tilde{\boldsymbol{m}}_r=\chi^2 \label{eq:diagonalconstraint}
\end{equation}
All matrices involved in Eq. \ref{eq:diagonalconstraint} are now diagonal, meaning the constraint can be reduced further to the following summation
\begin{equation}
    \sum_{i=1}^{3}\tilde{m}_{r,i}^2\frac{\sigma_i}{(\sigma_i+\lambda)^2}=\chi^2
\end{equation}
where $\sigma_i$ is the $i$th element of $\boldsymbol{\Sigma}_{rr}$

This final form is monotonic in $\lambda$ and can be solved reliably with a simple Newton's method implementation.
\begin{align}
    f(\lambda_n) &= \sum_{i=1}^{3}\tilde{m}_{r,i}^2\frac{\sigma_i}{(\sigma_i+\lambda_n)^2} - \chi^2 \\
    f'(\lambda_n)&=-\sum_{i=1}^{3}\tilde{m}_{r,i}^2\frac{2\sigma_i}{(\sigma_i+\lambda_n)^3}\\
    \lambda_{n+1}&= \lambda_n - \frac{f(\lambda_n)}{f'(\lambda_{n})}
\end{align}
Iterate until $||\lambda_{n+1} - \lambda_n||<\epsilon_\lambda$ and/or $||f(\lambda_n)||<\epsilon_f$. The final value $\lambda^*$ can then be substituted back into Eq. \ref{eq:subsol} to find $\boldsymbol{r}^*$. To close the loop on instantaneous safety, the following inequality is verified.
\begin{equation}
    ||\boldsymbol{r}^*||>r_k
\end{equation}

\section*{Appendix: Conic Form of Ellipsoid Projection Constraints}
All convex guidance programs introduced above include ellipsoid projection constraints to constrain uncertainty for passive safety. That type of constraint, given the reduced maneuver execution error model, is now explicitly formulated in convex form. The first approach program constraint is used as an example, with all projection constraints following the same structure. That constraint is
\begin{equation}
    r_k\leq(\hat{\boldsymbol{r}}_p^\mathcal{S})^\text{T}\,\delta\boldsymbol{m}_r^{\mathcal{S}}(t_f) - \chi\sqrt{(\hat{\boldsymbol{r}}_p^\mathcal{S})^\text{T}\delta\boldsymbol{P}_{rr}^{\mathcal{S}}(t_f)(\hat{\boldsymbol{r}}_p^\mathcal{S})}
\end{equation}
where $\delta\boldsymbol{m}_r^{\mathcal{S}}(t_f)$ and $\boldsymbol{P}_{rr}^{\mathcal{S}}(t_f)$ are both functions of the problem's decision variables $\Delta\boldsymbol{V}(t_0)$.

Let the cumulative state map be
\begin{equation}
    \boldsymbol{Q}=\boldsymbol{Q}_\mathcal{I}^\mathcal{S}(t_f,t_0)=\mathcal{R}_\mathcal{I}^\mathcal{S}(t_f)\boldsymbol{\Phi}(t_f,t_0)
\end{equation}
then the final position covariance post maneuver can be written as
\begin{equation}
    \boldsymbol{P}_{rr}^\mathcal{S}(t_f) = \boldsymbol{P}_{rr}^\mathcal{S}(t_f)_\text{fix} + \boldsymbol{Q}_{rv}\left[\sigma_{r,m}^2[\Delta\boldsymbol{V}(t_0)][\Delta\boldsymbol{V}(t_0)]^\text{T}+\sigma_{r,o}^2[\Delta\boldsymbol{V}(t_0)]_\times[\Delta\boldsymbol{V}(t_0)]_\times^\text{T}\right]\boldsymbol{Q}_{rv}^\text{T}
\end{equation}
\begin{equation}
    \boldsymbol{P}_{rr}^\mathcal{S}(t_f)_\text{fix} = \boldsymbol{Q}_{rr}\boldsymbol{P}_{rr}(t_0)\boldsymbol{Q}_{rr}^\text{T} + \boldsymbol{Q}_{rv}\boldsymbol{P}_{vr}(t_0)\boldsymbol{Q}_{rr}^\text{T} + \boldsymbol{Q}_{rr}\boldsymbol{P}_{rv}(t_0)\boldsymbol{Q}_{rv}^\text{T} + \boldsymbol{Q}_{rv}\left[\boldsymbol{P}_{vv}^{-}(t_0)+\sigma_a^2\mathbb{I}_{3}\right]\boldsymbol{Q}_{rv}^\text{T}
\end{equation}
When the covariance matrix is projected down to one dimension, the outer product term collapses, and the cross product operator can be redistributed to the projecting vector $\boldsymbol{u}$ such that
\begin{gather}
    \boldsymbol{w} =
    \begin{bmatrix}
        \sigma_{r,o}\left[\boldsymbol{u}\right]_\times\Delta\boldsymbol{V}(t_0) \\
        \sigma_{r,m}\boldsymbol{u}^\text{T}\Delta\boldsymbol{V}(t_0) \\
        \sqrt{(\hat{\boldsymbol{r}}_p^\mathcal{S})^\text{T}\boldsymbol{P}_{rr,\text{fix}}^\mathcal{S}(t_f)\hat{\boldsymbol{r}}_p^\mathcal{S}}
    \end{bmatrix}\in\mathbb{R}^5,\quad \boldsymbol{u}=\boldsymbol{Q}_{rv}^\text{T}\hat{\boldsymbol{r}}_p^\mathcal{S}\\
    ||\boldsymbol{w}|| = \sqrt{(\hat{\boldsymbol{r}}_p^\mathcal{S})^\text{T}\delta\boldsymbol{P}_{rr}^{\mathcal{S}}(t_f)(\hat{\boldsymbol{r}}_p^\mathcal{S})}
\end{gather}
With that restructuring, the ellipsoid projection constraint can be formulated in conic form as
\begin{gather}
    ||\boldsymbol{D}\Delta\boldsymbol{V}(t_0) + \boldsymbol{e}||\leq\boldsymbol{f}^\text{T}\Delta\boldsymbol{V}(t_0)+g \\
    \boldsymbol{D}=
    \begin{bmatrix}
        \sigma_{r,o}\left[\boldsymbol{u}\right]_\times \\
        \sigma_{r,m}\boldsymbol{u}^\text{T} \\ 0
    \end{bmatrix},\quad
    \boldsymbol{e}=
    \begin{bmatrix}
        \boldsymbol{0}_3 \\ 0 \\ \sqrt{(\hat{\boldsymbol{r}}_p^\mathcal{S})^\text{T}\boldsymbol{P}_{rr,\text{fix}}^\mathcal{S}(t_f)\hat{\boldsymbol{r}}_p^\mathcal{S}}
    \end{bmatrix}\nonumber\\
    \boldsymbol{f}=\boldsymbol{u}/\chi,\quad g=\left[(\hat{\boldsymbol{r}}_p^\mathcal{S})^\text{T}\left[\boldsymbol{Q}_{rr}\delta\boldsymbol{m}_r(t_0)+\boldsymbol{Q}_{rv}\delta\boldsymbol{m}_v(t_0)\right]-r_k\right]/\chi\nonumber
\end{gather}

%% file: references.bib
@inproceedings{ECOS,
  title={ECOS: An SOCP solver for embedded systems},
  author={Domahidi, Alexander and Chu, Eric and Boyd, Stephen},
  booktitle={2013 European control conference (ECC)},
  pages={3071--3076},
  year={2013},
  organization={IEEE}
}

@inproceedings{plaks2026closed,
  title={End-to-End Closed-Loop Rendezvous and Proximity Operations Performance for the CAPSTONE 02 Mission},
  author={Plaks, Connor and Down, Ian M. and Bolliger, Matthew and Bradley, Nicholas and Caudill, Michael},
  booktitle={2026 AAS/AIAA Astrodynamics Specialist Conference},
  pages={1--20},
  year={2026}
}

@book{rockafellar1997convex,
  title={Convex analysis},
  author={Rockafellar, R Tyrrell},
  volume={28},
  year={1997},
  publisher={Princeton university press}
}

@article{goulart2024clarabel,
  title={Clarabel: An interior-point solver for conic programs with quadratic objectives},
  author={Goulart, Paul J and Chen, Yuwen},
  journal={arXiv preprint arXiv:2405.12762},
  year={2024}
}

@book{boyd2004convex,
  title={Convex optimization},
  author={Boyd, Stephen and Vandenberghe, Lieven},
  year={2004},
  publisher={Cambridge university press}
}

@techreport{gates1963simplified,
  title={A simplified model of midcourse maneuver execution errors},
  author={Gates, Clarence R},
  year={1963},
  institution={NASA}
}

@article{malyuta2022convex,
  title={Convex optimization for trajectory generation: A tutorial on generating dynamically feasible trajectories reliably and efficiently},
  author={Malyuta, Danylo and Reynolds, Taylor P and Szmuk, Michael and Lew, Thomas and Bonalli, Riccardo and Pavone, Marco and A{\c{c}}{\i}kme{\c{s}}e, Beh{\c{c}}et},
  journal={IEEE Control Systems Magazine},
  volume={42},
  number={5},
  pages={40--113},
  year={2022},
  publisher={IEEE}
}

@article{morgan2014model,
  title={Model predictive control of swarms of spacecraft using sequential convex programming},
  author={Morgan, Daniel and Chung, Soon-Jo and Hadaegh, Fred Y},
  journal={Journal of Guidance, Control, and Dynamics},
  volume={37},
  number={6},
  pages={1725--1740},
  year={2014},
  publisher={American Institute of Aeronautics and Astronautics}
}

@article{berning2024chance,
  title={Chance-constrained, drift-safe guidance for spacecraft rendezvous},
  author={Berning Jr, Andrew W and Burnett, Ethan R and Bieniawski, Stefan},
  journal={arXiv preprint arXiv:2401.11077},
  year={2024}
}

@article{elango2025successive,
  title={Successive convexification for passively-safe spacecraft rendezvous on near rectilinear halo orbit},
  author={Elango, Purnanand and Vinod, Abraham P and Kitamura, Kenji and A{\c{c}}{\i}kme{\c{s}}e, Beh{\c{c}}et and Di Cairano, Stefano and Weiss, Avishai},
  journal={arXiv preprint arXiv:2505.17251},
  year={2025}
}

@inproceedings{elango2026continuous,
  title={Continuous-Time Successive Convexification for Passively-Safe Spacecraft Rendezvous on a Near Rectilinear Halo Orbit},
  author={Elango, Purnanand and Vinod, Abraham P and Kitamura, Kenji and Acikmese, Behcet and Di Cairano, Stefano and Weiss, Avishai},
  booktitle={AIAA SCITECH 2026 Forum},
  pages={2446},
  year={2026}
}

@incollection{gerstenmaier2019international,
  title={International rendezvous system interoperability standards (irsis)},
  author={Gerstenmaier, William H and Parker, David and Leclerc, Gilles and Shirama, Ryuichiro},
  booktitle={Technical Report},
  year={2019},
  publisher={Technical report, NASA}
}

@article{down2024autonomous,
  title={Autonomous Satellite Servicing Infrastructure for In-Space Assembly and Manufacturing},
  author={Down, Ian M and van Wijk, David EJ and Parikh, Deep and Majji, Manoranjan},
  journal={Journal of Manufacturing Science and Engineering},
  volume={146},
  number={12},
  pages={121008},
  year={2024},
  publisher={American Society of Mechanical Engineers}
}

@incollection{cheetham2021cislunar,
  title={Cislunar autonomous positioning system technology operations and navigation experiment (Capstone)},
  author={Cheetham, Bradley},
  booktitle={ASCEND 2021},
  pages={4128},
  year={2021}
}

@article{gardner2021capstone,
  title={Capstone: A cubesat pathfinder for the lunar gateway ecosystem},
  author={Gardner, Thomas and Cheetham, Brad and Forsman, Alec and Meek, Cameron and Kayser, Ethan and Parker, Jeff and Thompson, Michael and Latchu, Tristan and Rogers, Rebecca and Bryant, Brennan and others},
  year={2021}
}

@article{sullivan2017comprehensive,
  title={Comprehensive survey and assessment of spacecraft relative motion dynamics models},
  author={Sullivan, Joshua and Grimberg, Sebastian and D’Amico, Simone},
  journal={Journal of Guidance, Control, and Dynamics},
  volume={40},
  number={8},
  pages={1837--1859},
  year={2017},
  publisher={American Institute of Aeronautics and Astronautics}
}

@article{vela2026application,
  title={Application of Fundamental Modal Solutions to Relative Dynamics in the Cislunar Environment},
  author={Vela, Claudio and Opromolla, Roberto and Fasano, Giancarmine and Schaub, Hanspeter},
  journal={Journal of Guidance, Control, and Dynamics},
  volume={49},
  number={2},
  pages={344--358},
  year={2026},
  publisher={American Institute of Aeronautics and Astronautics}
}

@article{franzini2019relative,
  title={Relative motion dynamics in the restricted three-body problem},
  author={Franzini, Giovanni and Innocenti, Mario},
  journal={Journal of Spacecraft and Rockets},
  volume={56},
  number={5},
  pages={1322--1337},
  year={2019},
  publisher={American Institute of Aeronautics and Astronautics}
}

@mastersthesis{khoury2020orbital,
  title={Orbital rendezvous and spacecraft loitering in the earth-moon system},
  author={Khoury, Fouad},
  year={2020},
  school={Purdue University}
}

@article{vavrina2019safe,
  title={Safe rendezvous trajectory design for the restore-l mission},
  author={Vavrina, Matthew A and Skelton, C Eugene and DeWeese, Keith D and Naasz, Bo J and Gaylor, David E and D’souza, Christopher},
  journal={Advances in the Astronautical Sciences},
  volume={168},
  number={3649-3668},
  pages={176},
  year={2019},
  publisher={Univelt, Inc. San Diego, CA}
}

@article{elliott2022describing,
  title={Describing relative motion near periodic orbits via local toroidal coordinates},
  author={Elliott, Ian and Bosanac, Natasha},
  journal={Celestial Mechanics and Dynamical Astronomy},
  volume={134},
  number={2},
  pages={19},
  year={2022},
  publisher={Springer}
}

@article{leith2023time,
  title={A time regularization scheme for spacecraft trajectories subject to multi-body gravity},
  author={Leith, James and Russell, Ryan P},
  journal={The Journal of the Astronautical Sciences},
  volume={70},
  number={2},
  pages={7},
  year={2023},
  publisher={Springer}
}

@article{kulik2025applications,
  title={Applications of induced tensor norms to guidance navigation and control},
  author={Kulik, Jackson and Ruth, Maximilian and Orton-Urbina, Cedric and Savransky, Dmitry},
  journal={Journal of Guidance, Control, and Dynamics},
  volume={48},
  number={10},
  pages={2180--2198},
  year={2025},
  publisher={American Institute of Aeronautics and Astronautics}
}

@inproceedings{davis2022orbit,
  title={Orbit maintenance burn details for spacecraft in a near rectilinear halo orbit},
  author={Davis, Diane C and Scheuerle, Stephen T and Williams, Dale A and Miguel, Frederick S and Zimovan-Spreen, Emily M and Howell, Kathleen C},
  booktitle={2022 AAS/AIAA Astrodynamics Specialists Conference},
  number={AAS 22-545},
  year={2022}
}
